\documentclass[lettersize,journal]{IEEEtran}
\usepackage{amsmath,amsfonts}
\usepackage{algorithmic}
\usepackage{algorithm}
\usepackage{array}
\usepackage[caption=false,font=normalsize,labelfont=sf,textfont=sf]{}
\usepackage{textcomp}
\usepackage{stfloats}
\usepackage{url}
\usepackage{verbatim}
\usepackage{graphicx}
\usepackage{cite}

\usepackage{tabularx}
\usepackage{color}
\newtheorem{lemma}{\bfseries{Lemma}}
\newtheorem{theorem}{\bfseries{Theorem}}

\newcommand{\myref}[1]{Equation (\ref{#1})}
\usepackage{enumerate}
\usepackage{bm}
\usepackage{multirow}
\usepackage{multicol}
\usepackage{subfigure}
\usepackage{mathtools,cuted}
\allowdisplaybreaks[4] 

\usepackage{soul, color, xcolor}
\definecolor{myColor}{rgb}{255,0,0}
\makeatletter
\newcommand*{\new}{\@ifnextchar\bgroup{\new@}{\color{myColor}}}
\newcommand*{\new@}[1]{{\textcolor{myColor}{#1}}}
\makeatother

\makeatletter
\def\@eqnnum{{\normalfont \color{red} (\theequation)}}
\makeatother

\usepackage{xcolor}
\usepackage{xpatch}
\makeatletter
\def\changeBibColor#1{%
	\in@{#1}{
	}
	\ifin@\color{red}\else\normalcolor\fi
}
\xpatchcmd\@bibitem
{\item}
{\changeBibColor{#1}\item}
\makeatother

\begin{document}
	
	\title{
		Analyzing the Expected Hitting Time of Evolutionary Computation-based Neural Architecture Search Algorithms
	}
	
	\author{Zeqiong Lv,
		Chao Qian,~\IEEEmembership{Senior Member,~IEEE,}
		Gary~G.~Yen,~\IEEEmembership{Fellow,~IEEE,} \\
		and Yanan Sun,~\IEEEmembership{Senior Member,~IEEE}
			\thanks{This work was supported by National Natural Science Foundation of China (No. 62276175 and No. 62276124). \emph{(Corresponding author: Yanan Sun.)}}
		\thanks{Zeqiong Lv and Yanan Sun are with the College of Computer Science, Sichuan University, Chengdu 610065, China (e-mails: zq\_lv@stu.scu.edu.cn; ysun@scu.edu.cn).}
		\thanks{Chao Qian is with National Key Laboratory for Novel Software Technology, and School of Artificial Intelligence, Nanjing University, Nanjing 210023, China (e-mail: qianc@nju.edu.cn).}
		\thanks{Gary G. Yen is with the School of Electrical and Computer Engineering, Oklahoma State University, Stillwater, OK 74078 USA (e-mail:gyen@okstate.edu).}
		\thanks{Digital Object Identifier or DOI: 10.1109/TETCI.2024.3377683}
		\thanks{\copyright 2024 IEEE. Personal use of this material is permitted. Permission from IEEE must be obtained for all other uses, in any current or future media, including reprinting/republishing this material for advertising or promotional purposes, creating new collective works, for resale or redistribution to servers or lists, or reuse of any copyrighted component of this work in other works.}
	}

        \markboth{\tiny This work has been submitted to the IEEE for possible publication. Copyright may be transferred without notice, after which this version may no longer be accessible.}
        {Shell \MakeLowercase{\textit{et al.}}: A Sample Article Using IEEEtran.cls for IEEE Journals}

	\maketitle
	


	\begin{abstract}
		Evolutionary computation-based neural architecture search (ENAS) is a popular technique for automating architecture design of deep neural networks. Despite its groundbreaking applications, there is no theoretical study for ENAS. The expected hitting time (EHT) is one of the most important theoretical issues, since it implies the average computational time complexity. This paper proposes a general method by integrating theory and experiment for estimating the EHT of ENAS algorithms, which includes common configuration, search space partition, transition probability estimation, population distribution fitting, and hitting time analysis. By exploiting the proposed method, we consider the ($\lambda$+$\lambda$)-ENAS algorithms with different mutation operators and estimate the lower bounds of the EHT. Furthermore, we study the EHT on the NAS-Bench-101 problem, and the results demonstrate the validity of the proposed method. To the best of our knowledge, this work is the first attempt to establish a theoretical foundation for ENAS algorithms.
	\end{abstract}
	
	\begin{IEEEkeywords}
		Neural architecture search (NAS), evolutionary computation-based NAS (ENAS), average computational time complexity, expected hitting time.
	\end{IEEEkeywords}
	
	\section{Introduction} \label{Sec:intro}
    \IEEEPARstart{M}{anually} designed deep neural networks (DNNs) have been showing promising performance in diverse real-world applications~\cite{he2016deep,li2021survey}. However, this design process often demands rich expertise in both DNNs and domain knowledge of the problems to be solved. This has motivated researchers to conduct neural architecture search (NAS) that can automatically design promising DNN architectures. Typically, NAS is a challenging optimization problem~\cite{elsken2019neural}.

    The optimization algorithms primarily employed to address NAS can be classified into three categories: reinforcement learning (RL), gradient descent, and evolutionary computation (EC). In the early stages, NAS often employed the RL algorithms~\cite{zoph2016neural,liu2018progressive}, where a controller is iteratively updated based on the performance of the architecture as the reward to search for a better architecture. In recent years, the gradient descend-based NAS algorithms~\cite{liu2018darts,xu2019pc,chu2020fair} have been proposed to improve search efficiency. They generally relax the discrete architecture space to continuous one by mixture model and utilize gradient-based optimization to derive the best architecture. EC is another alternative for solving NAS~\cite{real2017large,liu2021survey}. Specifically, the EC is a family of heuristic algorithms that simulate the evolution of species or the behaviors of the population in nature, and the evolutionary algorithms (EAs) and swarm intelligence are the popular EC methods. In 2017, Google company took the initial step in proposing an ENAS algorithm~\cite{real2017large}. Since then, a slew of excellent ENAS algorithms has been proposed~\cite{sun2019completely,real2019regularized,xue2021self,lin2022evolutionary}, showing superior performance mainly in image classifications.

    Despite their success, there is still a significant lack of research on the theory of ENAS. The theory behind ENAS can help to understand how they work and why they are so successful while providing theoretical insights to enhance the ENAS methods. Among various theoretical issues, the expected first hitting time (EHT), which represents the average number of generations (iterations) needed to find an optimal solution~\cite{he2016average}, has been widely investigated in the EC community. Typically, the analysis of EHT involves calculating its upper and lower bounds. The upper bound of EHT signifies the worst-case time required by the algorithm. In practice, the immense computational resource budget derived from the EHT upper bound may not be easily accessible to every researcher. On the other hand, the lower bound of EHT represents the minimum time required by the algorithm. It can provide a minimum iteration for users with limited computational resources. Based on this practical significance, this study focuses on the lower bound of EHT as the start of EHT research on ENAS.

    In the past two decades, much literature has been devoted to the EHT analysis~\cite{doerr2019theory,zhou2019evolutionary}. Among those, representatives are the fitness-level method~\cite{wegener2003methods,lehre2011fitness,lassig2014general,doerr2021lower}, the convergence-based method~\cite{yu2008new}, and drift analysis method~\cite{he2001drift,doerr2012multiplicative,he2016average}. Similarly, these methods first analyze the state changes after one generation of the algorithm, and then substitute the analysis into the corresponding EHT theorems for obtaining EHT. However, the state change items analyzed by them are different, which include the probability of jumping into better states, the success probability of jumping into a target state, and the expected one-step progress. Taking the ONEMAX problem~\cite{droste2002analysis} for example, which aims to maximize the count of 1-bits in a binary-encoded string of length $n$, the fitness-level method analyzes the probability that each state increases at least one 1-bit (i.e., jumping into better states with more 1-bits). The convergence-based method analyzes the success probability of increasing the number of 1-bits of the current solution to $n$ (i.e., jumping into a target state with $n$ 1-bits). The drift analysis analyzes the expected number of the 1-bit increments (i.e., the expected progress in increasing 1-bits by one step). Switch analysis~\cite{yu2014switch,yu2015running} is another method for analyzing EHT, which derives the EHT of an algorithm by comparing it with a simpler algorithm. To be specific, it calculates the EHT difference of two algorithms by accumulating their one-step differences.

    Furthermore, these methods mostly focus on benchmark problems, such as the LEADINGONES problem~\cite{1997Convergence} and the TRAPZEROS problem~\cite{chen2012large}, in addition to the ONEMAX problem. This is because the benchmark problems typically have explicit fitness functions, which are used to analyze the state changes when calculating EHT. However, NAS typically has no explicit fitness functions~\cite{liu2021survey}. Given this, exploring the EHT in a purely theoretical manner is impractical. Statistical methods can overcome this issue by using samplings to simulate the key components in calculating the state change. In the literature~\cite{huang2019experimental}, an experimental method can model an average gain and then estimate the EHT. This insight inspired us to integrate statistical methods into existing analysis methods to analyze the EHT of ENAS algorithms. The contributions of our work are summarized as follows.
    \begin{enumerate}[1)]
        \item
        We define a combination encoding method as a common configuration to bridge the gap in modeling ENAS algorithms by the Markov chains. This work solves the problem that the variables of the ENAS algorithm are hard to be theoretically analyzed with the widely used binary encoding method.
        \item
        We employ the surface fitting method to model the population distribution function. This work introduces the experimental techniques into the evaluation of state changes, thereby establishing a connection between the EHT analysis of ENAS algorithms and NAS problems. 
        \item
        We consider the ($\lambda$+$\lambda$)-ENAS algorithms with the one-bit, $q$-bit, and bitwise mutation operators, and derive four theorems about the lower bounds of the EHT. This work presents the first mathematical expression of EHT for ENAS algorithms.
        \item
        We investigate the NAS-Bench-101 problem for case study and derive the EHT results of ENAS algorithms. The estimated EHT lower bounds are found to be lower than the experimental running time, providing evidence for the validity of the proposed method.
    \end{enumerate}
    
    The rest of the paper is organized as follows. In Section \uppercase\expandafter{\romannumeral2}, the preliminaries are provided. In Section \uppercase\expandafter{\romannumeral3}, the proposed method for estimating EHT is detailed. In Section \uppercase\expandafter{\romannumeral4}, the lower bounds of the EHT for ($\lambda$+$\lambda$)-ENAS algorithms using various mutation operators are investigated. In Section \uppercase\expandafter{\romannumeral5}, a case study about the proposed method is presented. In Section \uppercase\expandafter{\romannumeral6}, this paper is concluded along with future research directions.

	\section{Preliminaries}
    
    In this section, we introduce the NAS search space, present the main steps of the ($\lambda$+$\lambda$)-ENAS algorithms based on EAs considered in this paper, and overview the popular tools available for theoretically analyzing EAs.

    {
        \color{black}
        \subsection{Search Space of NAS}
        
        The search space of NAS defines the potential range of all architectures, similar to the role of decision space of general optimization problems. The NAS problem varies as the search space changes, leading to different optimal solutions. Here, we provide a brief overview of existing search space by using CNN as an illustrative example.
        
        Generally, the search space of NAS can be categorized into three groups based on basic units~\cite{liu2021survey}, which are layer-based, blocked-based, and cell-based. Fig.~\ref{fig:search_space} shows three architectures from these three search spaces. The layer-based search space denotes that the basic units are primitive layers, such as convolution layers, pooling layers, and fully-connected layers, as illustrated in Fig.~\ref{fig:search_space_a}. In the block-based search space, the basic units are predefined blocks consisting of fixed primitive layers and predefined topological relationships. Specifically, layers in the blocks have a specific topological relationship, such as the residual connection of ResBlock~\cite{he2016deep} and the dense connections of DenseBlock~\cite{huang2017densely}. Fig.~\ref{fig:search_space_b} is an example from the block-based search space based on ResBlock and DenseBlock. The cell-based search space is similar to the block-based, but its basic units are the same cells whose structures are determined during the search. Fig.~\ref{fig:search_space_c} is an architecture in the NAS-Bench-101 search space~\cite{ying2019bench} which is cell-based. In subsection~\ref{subsection:common_configuration}, we will introduce an encoding method applicable to these search spaces.

        \begin{figure}[!t]
            \centering
            \subfigure[]{
                \includegraphics[ height=4.5cm]{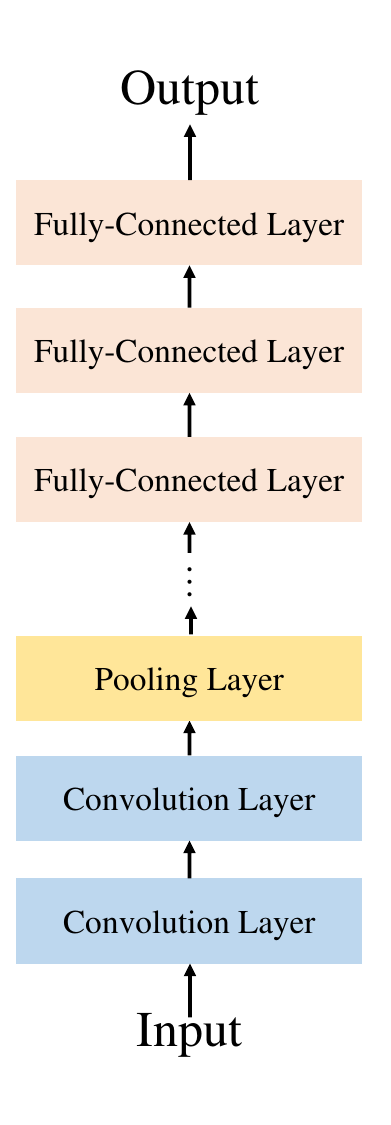}
                \label{fig:search_space_a}
            } 
            \subfigure[]{
                \includegraphics[ height=4.5cm]{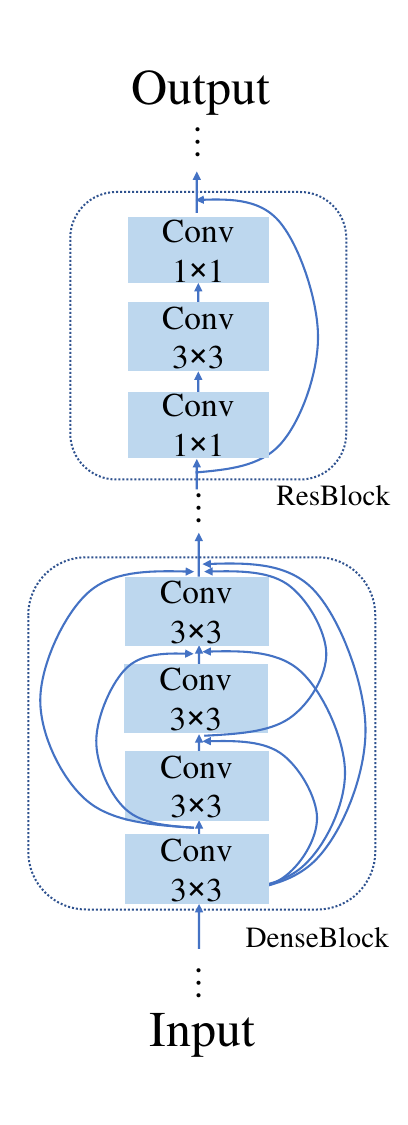}
                \label{fig:search_space_b}
            } 
            \subfigure[]{
                \includegraphics[ height=4.5cm]{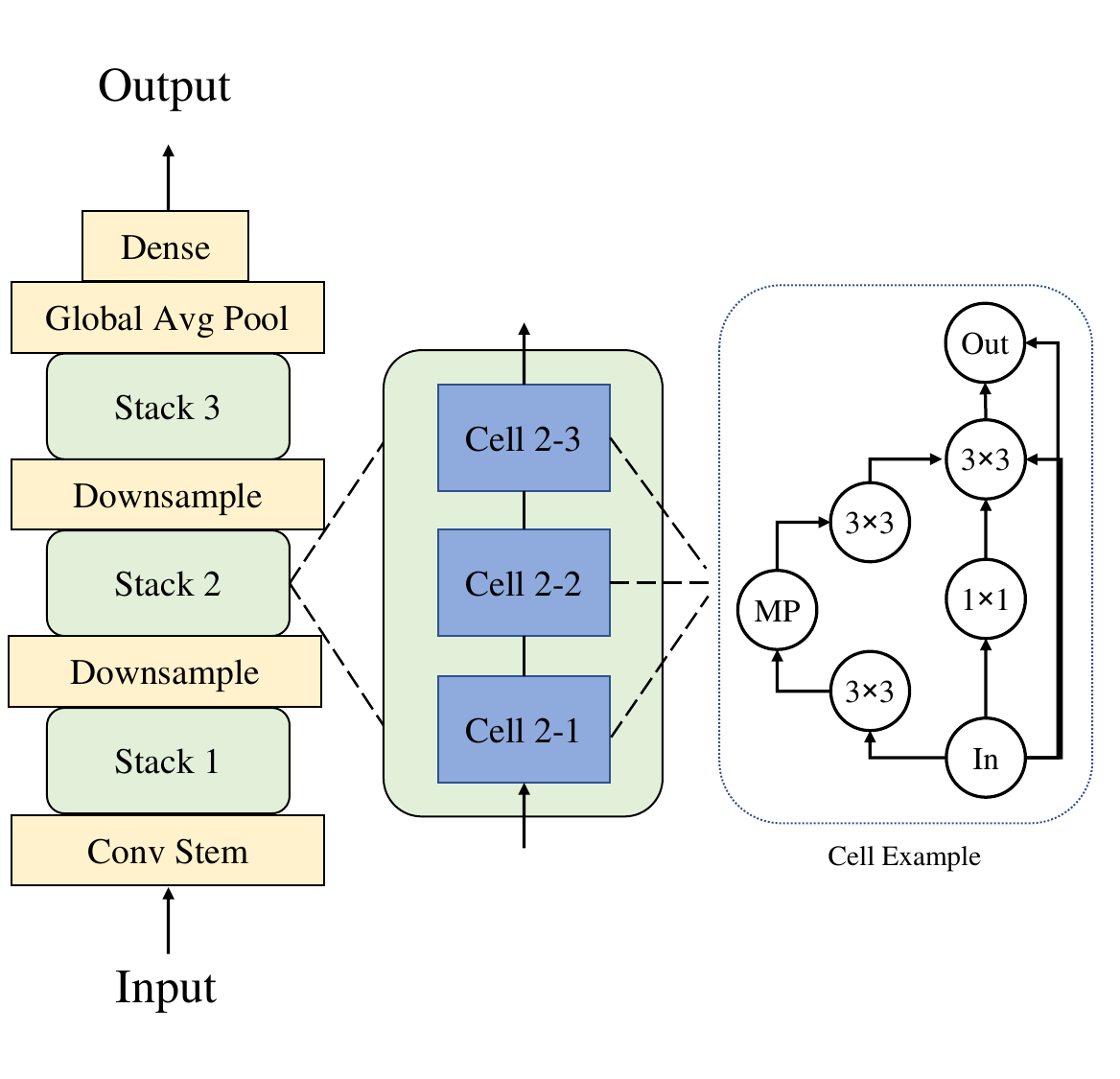}
                \label{fig:search_space_c}
            } 
            \caption{Three architectures for three kinds of search spaces, i.e., (a) an architecture in layer-based search space, (b) an architecture in block-based search space, and (c) an architecture in cell-based search space.}
            \label{fig:search_space}
            \vspace{-1em}
        \end{figure}

    }
    
    \subsection{Mutation-Based ENAS Algorithms}
    A commonly used EA for NAS is the ($\mu$+$\lambda$)-EA, which creates $\lambda$ offspring individuals in each generation and chooses the best $\mu$ individuals to survive into the next generation. There have been various variants of EAs that may result in different theoretical observations. Without loss of generality, in this work, we focus on the ($\lambda$+$\lambda$)-ENAS algorithms\footnote{We will treat ENAS as ($\lambda$+$\lambda$)-ENAS in the following discussions for the reason of brevity, unless specifying.}, based on ($\lambda$+$\lambda$)-EA, which is a general-purpose EA with a mutation only.

    Mutation operator is the most commonly used genetic operator in EAs and is widely studied in theory. Additionally, the elite selection strategy is often employed to ensure convergence. In the proposed method, we consider the ENAS algorithms that utilize the following commonly used mutation operators and selection strategies.
    \begin{itemize}
        \item \textbf{One-bit mutation}: Randomly flip one bit of solution.
        \item \textbf{$\bm{q}$-bit mutation}: Randomly flip $q$ bits of solution.
        \item \textbf{Bitwise mutation}: Uniformly flip each bit of each solution with mutation probability $p_m$. 
        \item \textbf{Truncation selection}: Select top $\lambda$ individuals from the current population ($2\lambda$) as offspring~\cite{chen2009new}.
        \item \textbf{Non-repeated selection}: Randomly select individuals one by one without repetition.
    \end{itemize}

    \subsection{Markov Chain Modeling}

	\begin{table}[!t]
        \caption{The mapping between core components of EA and Markov chain used in this study.\label{tab:notation_definitions}}
        \centering
        \begin{tabular}{cll}
            \hline
            Notations & EA & Markov Chain\\
            \hline
            \multicolumn{1}{c}{$s$} & \multicolumn{1}{l}{Solution / Individual} & \multirow{2}{*}{State} \\
            ${\xi_t}$ & Population \\
            \multicolumn{1}{c}{$S$} & \multicolumn{1}{l}{Individual Space} & \multirow{2}{*}{State Space} \\
            ${\chi}$ & Population Space \\
            $\chi^*$ & Target Population Space & Target State Space\\
            \hline
        \end{tabular}
        \vspace{-2em}
    \end{table}

    EAs evolve solutions from generation to generation, where the state of the next generation depends on the previous one. Thus, EAs can be modeled by Markov chains~\cite{he2003towards}. For clarity, the mapping between the core components of EAs and the Markov chains is shown in Table \ref{tab:notation_definitions}. Note that, we call a population optimal if it contains at least one optimal solution. The subspace composed of the optimal population is called target population space. 
	
	The EHT of the Markov chain can then be represented by~\myref{eq:secII_ET}:
    \vspace{-0.7em}
    \begin{equation} 
        \label{eq:secII_ET}
        {E}[T] =
        \sum_{t=0}^{+\infty} P(\xi_t \in \chi-\chi^*)
        \vspace{-0.5em}
    \end{equation}
    where $T=\min\{t\geq 0; \xi_t \in \chi^*|\xi_0\}$ represents the first hitting time of EAs starting with the initial state $\xi_0$, and $P(\xi_t \in \chi-\chi^*)$ is the probability that the $t$-th state is not optimal.

    \subsection{The Drift Analysis Approach} \label{Subsection:drift}
    Among the EHT analysis approaches introduced in Section~\ref{Sec:intro}, the fitness-level method usually relies on knowing the explicit form of the fitness function. However, the fitness value of state (i.e., architecture) in the NAS problem is usually derived by training the state on GPU with a dataset, which cannot be expressed through an explicit function. Furthermore, the convergence-based method is unsuitable for ENAS algorithms with one-bit mutation since they cannot successfully jump to the optimal solution in one step.
    
    In this paper, we use the drift analysis, in particular the average drift analysis~\cite{he2016average}, to estimate the EHT of the ENAS algorithms. Specifically, the key step in the drift analysis method is obtaining the one-step progress (drift). The standard drift analysis considers the upper/lower bound on the one-step change for all states, while the average drift analysis can tighten the bounds by considering the expected drift of all states. In practice, there are three steps for using the average drift analysis method: 1) identifying a distance function $d(\cdot)$ that measures the disparity between the state and the optimal point; 2) representing the average drift $\bar {\Delta}_t$~\cite{jagerskupper2008blend} by \myref{eq:average_drift} and the state distribution probability $\pi_t$; 3) using Lemma~\ref{Bound_EHT} to calculate EHT.

    \begin{equation}
        \label{eq:average_drift}
        \begin{aligned}
            \bar {\Delta}_t  
            = &
            \mathbb{E}[\mathbb{E}[d(\xi_t)-d(\xi_{t+1})\mid \xi_t=X\sim \pi_t]\mid X\notin\chi^* ]
        \end{aligned}
    \end{equation}

    \begin{lemma}[\cite{he2016average}]
        Given a Markov chain $\left\lbrace {\xi_t}\right\rbrace _{t=0}^{+\infty}$ converges to $\chi^*$ where initial state $\xi_0$ satisfies $P(\xi_0\in \chi-\chi^*)>0$, and an initial distance $d(\xi_0)$, for each generation $t$, there is an average drift $\bar{\Delta}_t$. 
        If $0<c_2 \leq \bar{\Delta}_t \leq c_1 $, where both $c_1$ and $c_2$ are greater than 0, then the EHT satisfies \myref{eq:aveEHT}:
        \begin{equation}\label{eq:aveEHT}
            \frac{\mathbb{E}[d(\xi_0)]}{c_1} \leq \mathbb{E}[T] \leq \frac{\mathbb{E}[d(\xi_0)]}{c_2}
            \vspace{-1em}
        \end{equation}
        \label{Bound_EHT}
    \end{lemma}
    where the expected initial distance $\mathbb{E}[d(\xi_0)]$ can be calculated by \myref{eq:ed0}:
    \begin{equation} \label{eq:ed0}
        \resizebox{1\width}{!}{ 
            $\mathbb{E}[d(\xi_0)]= \sum_{X \in {\chi}} P(\xi_0 = X)d(X)$
        }
        \vspace{-0.5em}
    \end{equation}

    In addition, several key issues need to be addressed before applying the average drift analysis method. These include establishing the fundamental configuration of encoding ENAS algorithms, addressing the challenges posed by partitioning the search space due to the intricate solution structure, analyzing the transition probability involved in standard drift calculation, and calculating the population distribution involved in average drift calculation. In this work, we will systematically propose a comprehensive method by addressing these issues for estimating the EHT of ENAS algorithms.

    \section{The Proposed Method}
	The framework of the proposed method, i.e., CEHT-ENAS, is shown in Algorithm \ref{framework}, which is composed of five parts: 
	\begin{algorithm}[h]
		\caption{The proposed CEHT-ENAS method}
		\label{framework}
		\textbf{Input}: ENAS algorithm with mutation operator, search space\\
		\textbf{Output}: EHT
		\begin{algorithmic}[1] 
			\STATE \textbf{\underline{Common configuration:}}  (subsection~\ref{subsection:common_configuration})
			\STATE\hspace{\algorithmicindent} $\left\lbrace {\xi_t}\right\rbrace _{t=0}^{+\infty}$ $\leftarrow$ Model the ENAS algorithm as a Markov chain process;
			\STATE\hspace{\algorithmicindent} Configure the parameters in the ENAS algorithm and search space;
			\label{alg_p1_start}
			\STATE\hspace{\algorithmicindent} $\{d(\cdot)\}$ $\leftarrow$ Define the distance function;\label{alg_p1_end}
			\STATE \textbf{\underline{Search space partition:}} (subsection~\ref{Subsection:search_space_partition})
			\STATE\hspace{\algorithmicindent} $\left\lbrace {\chi_i}\right\rbrace _{i=0}^{n}$ $\leftarrow$ Partition the search space into multiple subspaces according to the distance function;\label{alg_p2} 
			\STATE \textbf{\underline{Transition probability analysis:}} (subsection~\ref{Subsection:TP_individuals})
			\STATE\hspace{\algorithmicindent} $P^m(x,y)$ $\leftarrow$ Estimate the mutation transition probability from individual $x$ to individual $y$; \label{alg_p3_start}
			\STATE \textbf{\underline{Population distribution fitting:}} (subsection~\ref{Subsection:fitting})
				\STATE\hspace {\algorithmicindent} $\pi(\chi_d)$ $\leftarrow$ Find the distribution probability bound of a population with distance $d$ by surface fitting. \label{alg_p4}
			\STATE \textbf{\underline{The hitting time analysis:}} (subsection~\ref{Subsection:drift})
			\STATE\hspace{\algorithmicindent} $\Delta(X)$ $\leftarrow$ Estimate the point-wise drift of $X$;\label{alg_p5_start}
			\STATE\hspace{\algorithmicindent} $\bar {\Delta}_t$ $\leftarrow$ Estimate the average drift at the $t$-th generation; 
			\STATE\hspace{\algorithmicindent} $P \left( \xi_0 = X \right) $  $\leftarrow$ Estimate the probability of $\xi_0$ belongs to the non-target state space;
			\STATE\hspace{\algorithmicindent} $\mathbb{E}[d(\xi_0)]$ $\leftarrow$ Calculate the expected distance of the initial solution $x_0$.
			\STATE \textbf{return} $\mathbb{E}[T]$ $\leftarrow$ $\mathbb{E}[d({\xi}_0)]/\bar {\Delta}_t $
			\label{alg_p5_end}
		\end{algorithmic}
	\end{algorithm}
	
	\begin{itemize}
		\item
		\textbf{Common Configuration} (lines~\ref{alg_p1_start}-\ref{alg_p1_end}): We propose a genetic encoding approach tailored specifically for DNN architectures in ENAS algorithms. This will bridge the gap in modeling ENAS algorithms by the Markov chain. 
		\item
		\textbf{Search Space Partition} (line~\ref{alg_p2}): We define the distance measurement function for the current state space to the target state space. Based on this, the search space can be partitioned, and the subsequent transition probability can be estimated. In subsection \ref{Subsection:search_space_partition}, we give two methods of search space partition.
		\item
		\textbf{Individuals Transition Probability Analysis} (line~\ref{alg_p3_start}): We develop lemmas for calculating the transition probability between two individuals with different encoding methods and mutation operators. These serve as the foundation to perform the drift analysis. 
		\item
		\textbf{Average Population Distribution Fitting} (line~\ref{alg_p4}): We explore the surface fitting techniques and the statistical data to fit the population distribution of ENAS algorithms with different population sizes. With the fitting results, we can further use the average drift method to get the progress related to the fitness of specific NAS problems.
		\item
		\textbf{The Hitting Time Analysis} (lines~\ref{alg_p5_start}-\ref{alg_p5_end}): We use the average drift method (introduced in subsection~\ref{Subsection:drift}) to calculate the EHT lower bound.	
	\end{itemize}

	\subsection{Common Configuration of ENAS Algorithms} \label{subsection:common_configuration}
	
	\begin{figure}[h]
		\centering
		\includegraphics[scale=0.48]{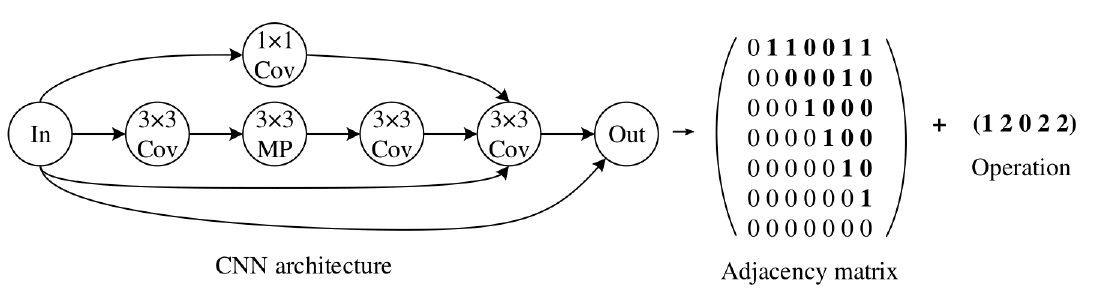}
		\caption{The architecture and encoding schematic of CNN.}
		\label{nasbench101}
		\vspace{-0.5em}
	\end{figure}
	
	The variables of ENAS algorithms refer to DNN architectures. Generally, a deep architecture is composed of two parts, one is the connection information of architecture, while the other is the operation information of the neural nodes. In the ENAS community, the binary string encoding method has been widely used to represent architectures. But this method makes itself unrealistic for theoretical analysis upon the variation between solutions. Specifically, the binary encoding method makes a significant difference between two different node operations, with more than one bit different between them (e.g., $0010$ represents a 3$\times$3 convolution operation, while $0100$ represents a 3$\times$3 maxpool operation, and two bits differ). As a result, a successful conversion between two operations would require mutating multiple bits simultaneously, which leads to a zero probability of conversion between partial operations via the one-bit mutation. This further results in certain node operations never appearing in candidate solutions when applying mutations, especially when the selection operator is involved. As a result, the binary string encoding method is not recommended for direct use in this work.
	
	In principle, any deep architecture can be formulated by directed cyclic graphs, where the nodes and edges of the graph correspond to the basic layers (or neuron operations) and connections between the layers. Motivated by this, we develop a combination genetic encoding method based on integers elaborately for ENAS in terms of its graph representation, which is composed of two parts. The first is the representation of the edges, which is composed of the upper triangular matrix, where ``1" represents an edge. The second comprises a string of integers $\left\lbrace 0,1,\ldots,L \right\rbrace$, where $L+1$ is the number of operation types of one layer. For clarity, we explain the method by using CNN as an example. As shown in Fig.~\ref{nasbench101}, the CNN architecture contains seven nodes (in, out, and five internal nodes) and nine edges, where each internal node represents one of the three operations (3$\times$3 convolution, 1$\times$1 convolution, and 3$\times$3 maxpool, i.e., $L=2$). Since the architecture is a directed cyclic graph, the elements of the 7$\times$7 upper triangular adjacency matrix describe the network connections. As a result, 21 bits can be used to represent the possible edges, which correspond to the encoding length of the first part. Similarly, the number of inter-nodes (i.e., the number is five in the example) corresponds to the length of the second encoding part. 
	
	It is worth noting that the combination encoding method can provide flexibility and accuracy in encoding and decoding architectures across different types of search spaces. In practice, the encoding methods can be simplified based on the characteristics of different search spaces. For the block-based and layer-based search spaces, where units are linearly arranged in a column-like structure, the encoding only needs to capture the changes in unit types and the number of units. The encoding of the topological structure is often unnecessary, and an integer encoding is typically sufficient. It is important to note that changing the number of units requires adjusting the length of the individual, where the crossover operator can effectively handle. For the cell-based search space, where architectures are composed of repeated blocks, a combination encoding mechanism can be employed to encode the repeated blocks and take it to represent the entire architecture. This method can significantly reduce the length of the encoding compared to encoding the entire architecture, while still preserving the essential information.
	
	To make the method more general to ENAS algorithms, the following common configuration is presented based on the above combination encoding method:
	\begin{itemize}
		\item Parameters: $v$ represents the maximum number of nodes in the search space of the network architectures, $n_1=v(v-1)/2$ represents the number of possible edges in the upper triangular matrix, $n_2=v-2$ refers to the number of intermediate vertices except the input and output vertices, $n=n_1+n_2$ refers to the individual length, and $\lambda$ specifies the population size.
		\item Encoding: Solution $s$ is encoded by combining string $a\in \{0,1\} ^{n_1} $ and string $b\in \{0,1,\ldots,L\}^{n_2}$, e.g., 
		\begin{quote}
			$s = (\underbrace{110011000101000100101}_{a=\{0,1\}^{21}}\circ \underbrace{12012}_{b=\{0,1,2\}^{5}})
			$
		\end{quote}
		\item Solution space: $S={\{0,1\}}^{n_1} {\{0,1,\ldots,L\}}^{n_2}$, and its size is $|S|=2^{n_1}(L+1)^{n_2}$.
		\item Population space: 
		$\chi = {\{\{0,1\}}^{n_1} {\{0,1,\ldots,L\}}^{n_2}\}^\lambda$, the population is considered as a disordered and repeatable set of solutions, so its size is $|\chi|={\lambda+|S|-1 \choose \lambda}$.
		\item Initialization: Randomly generate a population of $\lambda$ solutions encoded by the above encoding method.
		\item Fitness: The fitness of solution $s$ is defined as $f(s)$; the fitness of population $X$ is defined as the maximal fitness of its individuals, i.e., $f(X) = \max \{ f(s):s \in X \} $.
		\item Optimal solution: For simplicity, we assume that the solution $s^*$ with the highest fitness value is unique. 
		\item Distance: The distance of solution $s$ to the optimal solution is defined as $d(s)$; the distance of population $X$ is defined as the minimal distance of its individuals, i.e., $d(X)=\min\left\lbrace d(s):s\in X \right\rbrace$, which measures the distance from $X$ to the target population space $\chi^*$.
		\item Stopping criterion: The algorithm halts once $s^*$ is found.
	\end{itemize}

	Note that the utilization of the maximum fitness and minimum distance of individuals in population $X$ can effectively distinguish whether the populations are optimal or not, which is a crucial criterion for algorithm termination. Therefore, the fitness of population $X$ is defined as the maximal fitness of its individuals, and the distance of population $X$ is defined as the minimal distance of its individuals.
	
	According to the combination encoding method, we further update the calculation of solution number for each distance class in the solution space $S$. In the traditional encoding method $\{0,1\}^n$, there are $n$ individuals belonging to the class with distance $d$ in $S$. However, with the combination encoding $\{0,1\}^{n_1}\{0,1,..., L\}^{n_2}$, we need to divide the solution encoding into two parts, with respective distance values denoted as $d_1$ and $d_2$, where $d_1+d_2=d$. Then, we can express that the number of individuals with distance $d$ in $S$ is $C(d)$, as shown in \myref{equ:C(d)}.
	\vspace{-0.5em}
	\begin{equation} 
		\begin{aligned}
			C(d) & = 
			\left\{
			\begin{aligned}
				& 
				\resizebox{1\width}{!}{ 
					$	\sum_{d_1 = 0}^{d} {L^{d_2} {n_1\choose d_1}{n_2\choose {d_2}}}$
				}
				&d<n_1
				\\ & 
				\resizebox{1\width}{!}{ 
					$	\sum_{d_1 = \max\{d-n+n_1, 0\}}^{n_1} L^{d_2} {n_1\choose d_1}{n_2\choose {d_2}}$
				}
				&d\geq n_1
			\end{aligned}
			\right.
		\end{aligned}
		\label{equ:C(d)}
	\end{equation}
	
	To address the specific mutation requirements brought by the two-part encoding mechanism of deep architecture, which involves deciding which part to mutate, we introduce two new mutation strategies: bit-based fair mutation and offspring-based fair mutation. The bit-based fair strategy ensures that each bit is chosen equally, while the offspring-based fair mutation strategy ensures that each potential offspring has an equal chance of being born. For individuals encoded by the binary strings, both strategies ensure fair bit flipping during mutation and same probabilities for each type of offspring being generated. However, in the case of the ENAS algorithm, where the individuals are encoded using the combination method, the ``bit-based fair mutation" strategy would result in varying chances of generating possible offspring. This is because the non-binary encoding part needs to consider which new bit (i.e., a number between 0 and $L$, excluding itself) is chosen to mutate rather than simply flipping the bit between 0 and 1 as binary encoding does. To address this challenge and maintain fairness, we analyze the scenario in which the algorithm utilizes the ``offspring-based fair mutation" strategy in the theoretical analysis. With the two strategies defined above, we consider the following mutation operators.
	\vspace{-0.3em}
	\begin{itemize}
		\item \textbf{Mutation\#1} (One-bit mutation with ``bit-based fair mutation'' strategy): For each individual $x$, uniformly select an integer $r\in [1,n]$, and then mutate the $r$-th bit. When $r\in (n_1,n]$, it needs to additionally select an integer $l\in [0,L]\setminus \{x_r\}$ randomly, where $x_r$ represents the value of the $r$-th bit.
		Although each bit in non-binary encoding individual is fairly selected, each possible offspring birth is unfair. This is due to the fact that each bit in the non-binary encoding part can produce $L$ times more variants than the binary encoding part.
		\item \textbf{Mutation\#2} (One-bit mutation with ``offspring-based fair mutation'' strategy): For each individual, randomly select an integer number $r\in [1,n_1+Ln_2]$ with probability $ 1/(n_1+Ln_2)$. If $r\in [1,n_1]$, then flip the $r$-th bit; else mutate the $(n_1+\lceil \frac{r-n_1}{L} \rceil)$-th bit into one of the number from 
		$[0,L] \setminus \{x_i\}$, where ${x_i}$ represents the encoding value corresponding to the $(n_1+\lceil \frac{r-n_1}{L} \rceil)$-th bit of individual $x$.
		In this way, the probability of selecting any bit belonging to the first $n_1$ bits is $1/(n_1+Ln_2)$, which is less than the probability of selecting from the last $n_2$ bits, i.e., $L/(n_1+Ln_2)$. This mutation strategy ensures that the individuals who cannot use the binary encoding method can mutate fairly. That is, each sub-generation has the same chance of being generated. This is not guaranteed by the traditional \textbf{Mutation\#1}.
		\item \textbf{Mutation\#3} ($q$-bit mutation with ``bit-based fair mutation'' strategy): For each individual, randomly select $q$ unique integers $\{q_i\}_{i=1}^n$ with probability $1/{n \choose q}$, where $q_i\in (0,n]$, and then mutate the $q$ bits. The mutation operation of each bit is similar to \textbf{Mutation\#1}. 
		\item \textbf{Mutation\#4} (bitwise mutation): 
		For each individual, independently mutate each bit with probability $p_m=1/n$.
	\end{itemize}

	\subsection{Search Space Partition} \label{Subsection:search_space_partition}
	
	The transition probability is the key to theoretically investigating ENAS algorithms. This is because it serves the the basis for understanding how the initial population progresses to the target population. In order to facilitate the analysis of transition probability, we intend to partition the search space into several subspaces according to some bases, e.g., fitness, distance, and so on. In the following, we first discuss the definition of the distance function and then consider the basis for partitioning the search space from two distinct aspects.
	
	Hamming distance, commonly utilized in EC theoretical analysis, counts the number of positions with different symbols between two strings (encodings). It serves to measure the progress made between two states and accurately characterizes the encoding changes before and after mutation. Therefore, Hamming distance is useful in analyzing individual differences, especially in understanding the transition probability of individual mutation based on the combination encoding method. In this paper, we employ this simple distance function to make a first attempt at the EHT analysis of ENAS algorithms. Specifically, we use the Hamming Distance to measure how far an individual $s$ is away from the optimal point $s^*$, i.e., the distance $d(s)$. However, the Hamming distance does not capture the differences in fitness values, such as classification accuracy in a classification task, between two individuals. Addressing this limitation would require constructing a distance function based on fitness, which is currently infeasible without NAS-specific fitness function expression. Therefore, when using the Hamming distance, we focus on the progress made in optimizing the encoding rather than optimizing fitness. Recognizing this limitation, we will incorporate experimental methods in subsequent analysis steps (Section~\ref{Subsection:fitting}) to establish the relationship between progress drift and fitness.
	
	Then, we discuss the basis for partitioning the search space. From the perspective of utilizing fitness value as the criterion to divide search space, we can directly define the optimal subspaces and the non-optimal subspaces. But for millions of solutions that often exist in an ENAS algorithm, we cannot exactly know how much the fitness value has changed after the individual has performed the mutation operation, and we cannot divide the search space with such vanilla approaches. Therefore, we consider the distance as the space partition. From the perspective of individuals, the search space which is also called solution space $S$, is a set of integer strings. According to the distance value, we divide $S$ into $(n+1)$ subspaces without overlaps, i.e., $\{S^i\}^n_{i=0}$. The distance values of all solutions in $S^i$ are $i$. For population space $\chi$, we partition it into $(n+1)$ subspaces $\{\chi_i\}^n_{i=0}$, where the subspace $\chi_0$ equals to $\chi^*$ that contains all the optimal populations (the distance of the optimal solution in the population is zero). All populations in the subspace $\chi_i(i\in\left\lbrace 1,2,\ldots,n\right\rbrace )$ are non-optimal solutions, and the minimum distance value of the solutions in each population is $i$. The space $\chi_i$ can be subdivided into $\lambda$ small subspaces, i.e., $\{\chi_i^{\gamma} \} ^n_{i=1} (\gamma=1 ,2, \dots,\lambda)$, where $\gamma$ represents the number of individuals of which the distance values are $i$. Then, there are $\lambda(n+1)$ subspaces. The size of subspace $\chi_i$ can be denoted by $|\chi_i| = \sum_{\gamma = 1}^{\lambda} |\chi_i^\gamma|$, where $|\chi_i^\gamma|$ can be calculated by~\myref{chi_i_r}:
	\vspace{-0.5em} 
	\begin{equation}
		\label{chi_i_r}
		\resizebox{1\width}{!}{ 
			$|\chi_i^\gamma| = {{\gamma + C(i) -1} \choose \gamma} {
				{
					{\lambda -\gamma + \sum_{j=i+1}^{n}C(j)} -1
				} \choose {\lambda-\gamma}
			}$
		}
		\vspace{-0.5em} 
	\end{equation}
	where $C(i)$ is the number of solutions with distance $i$. With these designs, we can make it easier to analyze the transition probability of the states in the two subspaces, which will be introduced next.

	\subsection{\color{black}Transition Probability Analysis} \label{Subsection:TP_individuals}
	
	In this subsection, we provide lemmas to enable the investigation of the transition probability between individuals. They are applicable to analyze other EC scenarios using the combination encoding method and provide insights for further studying evolution operators. Specifically, we introduce Lemmas~\ref{lemma:combination_q_tp_onebit_M1} and \ref{lemma:combination_q_tp_onebit_M2} to depict the transition probabilities using one-bit mutation (i.e., \textbf{Mutation\#1} and \textbf{Mutation\#2}), and Lemmas~\ref{lemma:combination_q_tp_onebit_M3} and \ref{lemma:combination_q_tp_onebit_M4} to depict the transition probabilities using $q$-bit mutation (\textbf{Mutation\#3}) and bitwise mutation (\textbf{Mutation\#4}). Note that, the lemmas are derived from elementary combinatorics and probability theory. Due to space limitations, we omit the detailed proofs in the supplementary material.
	
	The following lemmas are all based on the definition that $d$ denotes the Hamming distance between a solution and a unique optimal solution. The distance of a combination-encoded individual $x$ with length $n$ is $d_x=d_1+d_2$, where $n_1=v(v-1)/2$ and $n_2=v-2$.
	\begin{lemma}
		After using \textbf{Mutation\#1}, $x$ can generate offspring $y$ with distance $d_y=d_x + i$ $(i\in[-1,1])$. There are three cases for the mutation transition probability $P_1^m$: 1) if $d_y=d_x-1$, it is $P_1^m=(d_1L+d_2)/{(nL)}$; 2) if $d_y=d_x$, it is ${d_2 (L-1)}/{(nL)}$; 3) otherwise, if $d_y=d_x+1$, it is $1 - d/n$.
		\label{lemma:combination_q_tp_onebit_M1}
	\end{lemma}
	
	\begin{lemma}
		After using \textbf{Mutation\#2}, $x$ can generate offspring $y$ with distance $d_y=d_x + i$ $(i\in[-1,1])$. There are three cases for the mutation transition probability $P_2^m$: 1) if $d_y=d_x-1$, the transition probability is given by ${d_x}/(n_1+Ln_2)$; 2) if $d_y=d_x$, it is given by ${d_2 (L-1)}/(n_1+Ln_2)$; 3) otherwise, if $d_y=d_x+1$, it is $1 - ({d_1 + L d_2})/(n_1+Ln_2)$.
		\label{lemma:combination_q_tp_onebit_M2}
	\end{lemma}

	\begin{lemma}
		After using \textbf{Mutation\#3}, $x$ can generate offspring $y$ with distance $d_y=d_x+q-j$ $(j\in[0,2q])$. The mutation transition probability $P_3^m$ can be calculated by \myref{eq:pmxy_M3}:
		\begin{equation}\label{eq:pmxy_M3}
			\begin{cases}
				P_3^m(x,y) = 
				\sum_{z=0}^{\min{\{q,n_2\}}}
				\sum_{a=0}^{\min{\{q-z,d_1\}}}
				\sum_{b=0}^{\min{\{z,d_2\}}}
				{
					W
				}/{
					\binom{n}{q}
				}
				\\
				W = 
				\binom{d_1}{a}
				\binom{n_1-d_1}{q-z-a}
				\binom{d_2}{b}
				\binom{n_2-d_2}{z-b}
				\binom{b}{c}
				\left({1}/{L}\right)^c
				\left(1-{1}/{L}\right)^{b-c}
				\\
				c=j-2a-b
			\end{cases}
		\end{equation}
		Note that since some combinations of $a$ and $b$ will not satisfy $j$, i.e., $c< 0$ or $c> b$, so at this point there is $\binom{b}{c}=0$.
		\label{lemma:combination_q_tp_onebit_M3}
	\end{lemma}
	
	\begin{lemma}
		After using \textbf{Mutation\#4}, $x$ can generate offspring $y$ with distance $d_y \in[0,n]$. The mutation transition probability $P_4^m$ can be calculated by \myref{eq:pmxy_M5}:
		\begin{equation} \label{eq:pmxy_M5}
			\begin{aligned}
				P_4^m(x,y) = & 
				\sum_{q=0}^{n}
				\binom{n}{q} 
				\left(\frac{1}{n}\right)^q \left(1-\frac{1}{n}\right)^{n-q}
				P_3^m(x,y)
			\end{aligned}
			\vspace{-0.5em}
		\end{equation} 
		where $P_3^m(x,y)$ is derived by~\myref{eq:pmxy_M3}.
		\label{lemma:combination_q_tp_onebit_M4}
	\end{lemma}

	So far, we have discussed the mutation transition probability of five cases, and our findings will serve as the core theoretical foundation for using drift to analyze running time.
	
	Since the utilization of elite selection strategy, the distribution probability $P^s$ of selection satisfies \myref{eq:psxyz}:
	\vspace{-0.3em} 
	\begin{equation} \label{eq:psxyz}
		P^s(x,y,z) = 
		\begin{cases}
			P(f(x)\geq f(y)), & {z=x}
			\\
			P(f(x) < f(y)), & {z=y}
		\end{cases}
		\vspace{-0.5em} 
	\end{equation}
	where $z$ represents the individual entering the next generation. Therefore, the transition probability from $x$ to $z$ can be represented as $P(x,z) =  P^m(x,y) \cdot P^s(x,y,z)$. Please note that the transition probability does not change with the transformation in $t$ because no self-adaptive strategy is employed.

	\subsection{Fitting Population Distribution} \label{Subsection:fitting}
		
	As discussed in Subsection~\ref{Subsection:search_space_partition}, the population distribution $\pi_t$ is obtained by experiments. Specifically, we use sampling data to fit distribution $\pi_t(\chi_d)$ which represents the probability of the current population belonging to $\chi_d$ at the $t$-th generation. We first detail the {steps for collecting sampling data:}
	\begin{enumerate}[1)]
		\item Define the mutation operation and selection operation, and specify the ranges for population size.
		\item Determine the problem size which also refers to the solution size $n$ in one specific search space. Consequently, $n+1$ potential values for individual distance exist.
		\item Determine the number of algorithm loops required for statistical experiments. For example, running each ENAS algorithm 1,000 times to derive statistical data.
		\item Construct $n+1$ tables to record which subspace the population belongs to during the algorithm executing process. In the $i$-th table, the horizontal axis represents the number of times the algorithm runs, and the vertical axis represents the generation number. Each item in the table records if the $t$-th generation population belongs to the subspace $\chi_i$ when the algorithm runs for the $j$-th time, and if so, then one is recorded in this table; otherwise zero.
		\item Execute the ENAS algorithm 1,000 times and complete the $n$ tables. Sum the values of the $t$-th row in $d$-th table, and divide by 1,000 to get the probability $\pi_t(\chi_d)$. Then, we can easily calculate a lower bound or upper bound of the probabilities $\{\pi_1(\chi_d),\pi_2(\chi_d),...\}$, i.e., $\pi_{lower}(\chi_d)$ or $\pi_{upper}(\chi_d)$.
		\item Change the population size and repeat the previous five steps until all the population sizes are adopted.
		\item Construct a 3D plot to visualize relationships among population sizes, population hamming distances, and population distribution probability bounds (lower or upper).
	\end{enumerate}
	
	Next, we transform the sampling data into the mathematical expression for further derivation about $\pi_t(\chi_d)$. Surface fitting methods are exactly suited for this task as they are designed to reconstruct the continuum from scattered data points by applying mathematical tools~\cite{lancaster1986curve}. After collecting experimental data, finding a suitable function $\pi_t$ is an important step in applying average drift (Lemma~\ref{Bound_EHT}) to analyze the EHT of ENAS algorithms. We use the surface fitting techniques to estimate $\pi_t$ by fitting a surface of population distribution with respect to the distance difference $d$ and population size $\lambda$. Based on the fitting surface, we then calculate the lower bound or upper bound of $\bar {\Delta}_t$ that meets the condition of Lemma~\ref{Bound_EHT}. 
	
	It is worth noting that since we aim to analyze the bound of EHT, the calculation of population distribution as a step in EHT analysis also needs to satisfy the boundary constraints. Therefore, the surface fitting method must be capable of addressing constrained surface fitting problems. Under the constraint conditions, the data points used for fitting are distributed either above or below the fitting surface, which depending on whether we are analyzing the upper bound or lower bound of EHT. Specifically, if we aim to calculate the lower bound of EHT (i.e., satisfying the condition $\bar{\Delta}_t \leq c_1$ of Lemma~\ref{Bound_EHT}), we need to further assess the correlation between $\bar{\Delta}_t \leq c_1$ and the population distribution $\pi_t$. If they have a positive correlation, then the upper bound of the population distribution needs to be found, which means we need to ensure that the data points are distributed below the fitted surface; conversely, we need to ensure that the data points are distributed above the fitted surface. In this paper, as the first attempt to analyze the EHT of the ENAS algorithm, we adopt a simple method, i.e., the least squares method, for fitting instead of more complex methods. Based on the results of surface fitting, Lemma~\ref{Bound_EHT} can be further applied to analyze the upper bound on the running time of ENAS algorithms.

	\section{Lower Bounds Analysis}
	
	In this section, we provide the mathematical expressions for the lower bounds of EHT for four ($\lambda$+$\lambda$)-ENAS algorithms using different mutation operators (\textbf{Mutation\#1-Mutation\#4}).
	
	\subsection{($\lambda$+$\lambda$)-ENAS algorithm with one-bit mutation}
	
	By using Lemmas~\ref{lemma:combination_q_tp_onebit_M1} and \ref{lemma:combination_q_tp_onebit_M2}, we present two theorems about the EHT of the ($\lambda$+$\lambda$)-ENAS algorithms for two types of one-bit mutation operators. 
	The first operator (i.e., Mutation\#1 operator) uses the one-bit mutation strategy that is fair to mutate each bit, and the second operator (i.e., Mutation\#2 operator) is used to ensure the fairness of the offspring.
	The following are the specific theorems and their proof process.

	\begin{theorem}
		For a ($\lambda$+$\lambda$)-ENAS algorithm using the one-bit mutation (\textbf{Mutation\#1}) and the non-repeated selection mechanism, its EHT is lower bounded by \myref{eq:lbEHT_mutation1}:
		\vspace{-0.3em} 
		\begin{equation}
			\frac{
				n (1-\pi_t(\chi^*))
				\sum_{d=1}^n d \pi_0{(\chi_d)}
			}{
				\sum_{d=1}^n d
				\sum_{\gamma=1}^\lambda \gamma \pi_t(\chi_d^\gamma)
			}
			\label{eq:lbEHT_mutation1}
			\vspace{-0.3em} 
		\end{equation}
		where $n$ is the solution size, $\lambda$ is the population size, $\pi_t{(\chi^*)}$ and $\pi_0{(\chi_d)}$ can be calculated by population distribution $\pi_t$.	
		\label{theorem:CEHTENAS_M1}
	\end{theorem}	
	
	\begin{IEEEproof}
        Based on CEHT-ENAS method, the process of ($\lambda$+$\lambda$)-ENAS algorithm is first modeled by
		$\left\lbrace {\xi_t}\right\rbrace _{t=0}^{+\infty} ({\xi_t}\in \chi)$. For the search space $\chi$, we partition it with two previously introduced methods:
		1) partition search space by the minimum distance value $i$ of solutions in each population, i.e., $ \left\lbrace \chi_i \right\rbrace ^n_{i=0}$, thus there are $(n+1)$ subspaces;
		2) subdivide the space $\chi_i$ into $\lambda$ subspaces, 
		i.e., $\{\chi_i^{\gamma} \} ^n_{i=0} (\gamma=1 ,2,\dots,\lambda)$,
		where $\gamma$ represents the number of individuals of which the distance values are $i$, thus there are $\lambda(n+1)$ subspaces. Next, the proof process mainly comprises four steps: 1) calculate the mutation transition probability; 2) calculate the point-wise drift; 3) calculate the average drift; 4) calculate the EHT.
		
		1) At the $t$-th generation, $\xi_t$ can be any population $X$ in the search space $\chi_k$, where $k>0$. During one generation, there are $\lambda$ new individuals generated which are denoted as population $Y$. It is known that the distance value of the population $X$ belongs to $[k, n]$, so the distance value of these $\lambda$ new individuals belongs to $[k-1, n]$. Following, we calculate the transition probability when the next generation $\xi_{t+1}$ belongs to any population $Y \in \chi_{k-1}$. The following is the specific analysis process.	
		Firstly, we calculate the probability that at least one individual in $Y$ is with distance value $k-1$, and denote it as $P(Y\in\chi_{k-1}|X)$.
		Then, the mutation transition probability from $X$ to $Y$ is estimated as $P(Y\in\chi_{k-1}|X)$.	
		Secondly, selecting $\lambda$ elite individuals from the populations $X$ and $Y$ as the population $Z$. We calculate the probability of individuals whose distance function value is $k-1$ in population $Z$, and denote it as $P(Z\in\chi_{k-1}\mid X,Y)$.
		Finally, the transition probability from $X$ to $Z\in\chi_{k-1}$ is estimated by $P(Y\in\chi_{k-1}|X)\cdot P(f(Y)>f(X))$.	
		
		Next, the calculation of mutation transition probability is detailed. Using non-repeated selection mechanism to select a population $X \in \chi_k$ as parents, the mutated population $Y$  belongs to $\chi_{k-1} \cup \chi_{k} \cup \chi_{k+1}$. As a result, the probability that $Y$ belongs to $\chi_{k-1}$ is represented as \myref{equ:E(P(Y|X))_mutation1}:
		\vspace{-0.6em}
		\begin{equation} 
			\begin{aligned}
				& P(Y \in \chi_{k-1} \mid X \in \chi_k) 
				\\ = & \mathbb{E} [P(Y \in \chi_{k-1} \mid X \in \chi_k^\gamma)]
				= 
				\mathbb{E}[1-(1-P^m(k,k-1))^\gamma] 
				\\ = &
				\sum_{\gamma = 1}^{\lambda}
				\left(1- 
				\prod^\gamma_{i=1} 
				{\left(1-\frac{(L d^i_1+ d^i_2)}{nL}\right)} \right)
				\frac{
					P(X\in\chi_k^\gamma)
				}{
					P(X\in\chi_k)
				}
				\\ \leq &
				\sum_{\gamma = 1}^{\lambda}
				\left(1-{\left(1-\frac{k}{n}\right)}^\gamma \right)
				\frac{
					P(X\in\chi_k^\gamma)
				}{
					P(X\in\chi_k)
				}
				\\ \leq &
				\frac{k}{n}
				\sum_{\gamma=1}^{\lambda}
				{
					\gamma
					\frac{
						P(X\in \chi_k^\gamma)
					}{P(X\in \chi_k)}
				}
				=
				\frac{k}{n}
				\sum_{\gamma=1}^{\lambda}
				{
					\gamma
					\frac{
						\pi_t(\chi_k^\gamma)
					}{\pi_t(\chi_k)}
				}
			\end{aligned}
			\label{equ:E(P(Y|X))_mutation1}
			\vspace{-0.2em}
		\end{equation}
		where $P^m(k,k-1)$ refers to probability of an individual with distance $k$ mutated to an individual with distance $(k-1)$ by using one-bit mutation (\textbf{Mutation\#1}) (can be obtained by Lemma~\ref{lemma:combination_q_tp_onebit_M1}), 
		$d^i_1$ and $d^i_2$ represent the corresponding values of $d_1$ and $d_2$, respectively, of the $i$-th solution with distance $k$,
		the probability distribution of random variable $\gamma$ is equal to the probability distribution that population $X$ belongs to $\chi_k^{\gamma} $, i.e., 
		$P(\gamma) = P(X \in \chi_k^{\gamma} \mid X\in\chi_k) 
		= P(X\in\chi_k^\lambda) / P(X\in\chi_k) 
		$, 
		the first inequality is holds by $(L d^i_1+d^i_2)/L\leq k$ for any solution with distance $k$, 
		and the second inequality can be easily derived from Bernoulli's inequality.
		
		2) Based on the above analysis, the point-wise drift can be calculated by \myref{con:drift_t_mutation1_without_d}:
		\begin{equation}
			\begin{aligned}
				\Delta(X) 
				= & 
				\mathbb{E}[d(\xi_t)-d(\xi_{t+1}) \mid \xi_t = X]
				\\ = &  
				\resizebox{1\width}{!}{  
					$\sum_{Z \in \chi}P(X,Z){(d_X-d_Z)} $
				}
				\\ = &
				\resizebox{1\width}{!}{  
					$\sum_{k = 0}^{n} \sum_{Z \in \chi_k}P(X,Z){(d_X-d_Z)} $
				}
				\\ = &
				\resizebox{1\width}{!}{  
					$\sum_{Z \in \chi_{[(d_X-1):n]}}
					P(X,Z){(d_X-d_Z)} $
				}
				\\ \leq &
				\resizebox{1\width}{!}{  
					$\sum_{Z \in \chi_{(d_X-1)}}P(X,Z){(d_X-(d_X-1))} $
				}
				\\ \leq &
				P(Y \in \chi_{(d_X-1)}\mid X)
			\end{aligned}
			\label{con:drift_t_mutation1_without_d}
		\end{equation}
		where $\chi_{[(d_X-1):n]}$ represents $\left\lbrace \chi_{d_X-1} \cup \chi_{d_X} \ldots \cup \chi_n \right\rbrace $,
		the first inequality is based on the consideration of the most optimistic situation that population $d(Z)=d(X)-1$, 
		and the second inequality is based on the most optimistic situation that population $Y$ is the same as $Z$.
		Thus, $\Delta(X \in \chi_d)$ can be calculated by \myref{con:drift_t_mutation1}:
		\vspace{-0.5em}
		\begin{equation}
			\begin{aligned}
				\Delta(X \in \chi_d) &\leq P(Y \in  \chi_{d-1} \mid X \in \chi_d ) &\leq 
				\frac{d}{n}
				\sum_{\gamma=1}^{\lambda}
				{
					\gamma
					\frac{
						\pi_t(\chi_{d}^\gamma)
					}{\pi_t(\chi_{d})}
				}
			\end{aligned}
			\label{con:drift_t_mutation1}
		\end{equation}
		where the last equation is derived from \myref{equ:E(P(Y|X))_mutation1}.
		It can be observed that \myref{con:drift_t_mutation1} is only related to $d_{X}$. 
		In the following, $\Delta(X \in \gamma_d)$ will be briefly denoted as $\Delta(d)$.

		3) Since the possible state of $X$ at $\xi_t$ is a random variable,
		we get the expectation of its function $\Delta(X)$, i.e., the average drift $\bar {\Delta}_t$, to indicate the progress of two neighbour generations. 
		Thus, the $\bar {\Delta}_t$ can be calculated by \myref{con:ave_drift_t_mutation1}:
		\vspace{-0.5em}
		\begin{equation} 
			\begin{aligned}
				\bar {\Delta}_t 
				= & 
				\mathbb{E}[\mathbb{E}[d(\xi_t)-d(\xi_{t+1})\mid \xi_t=X\sim \pi_t]\mid X\in\chi-\chi^* ]
				\\ = & 
				\resizebox{1\width}{!}{  
					$\sum_{X \in \chi-\chi_*}
					{\Delta (X) P(\xi_t=X)}/{P(\xi_t \in {\chi-\chi^*})}	 $
				}
				\\ = &
				\resizebox{1\width}{!}{  
					$ 
					\sum_{d=1}^n \Delta{(d)}P(d(\xi_t)=d)
					/(1-\pi_t{(\chi^*)})$
				}
				\\ = &
				\resizebox{1\width}{!}{  
					$ 
					\sum_{d=1}^n \Delta{(d)} {\pi_t{(\chi_d)}} /(1-\pi_t{(\chi^*)})$
				}
				\\ \leq &
				\frac{1}{n (1-\pi_t{(\chi^*)})}
				\sum_{d=1}^n 
				d
				\sum_{\gamma=1}^{\lambda}
				{
					\gamma
					\pi_t(\chi_{d}^\gamma)
				}
			\end{aligned}%
			\label{con:ave_drift_t_mutation1}
			\vspace{-0.5em}
		\end{equation}
		
		4) At the initial moment, the expectation of $d(\xi_0)$ can be calculated by \myref{equ:Ed0_mutation1}:
		\vspace{-0.5em}
		\begin{equation}
			\begin{aligned} 
				\mathbb{E}[d(\xi_0)]= & 
				\resizebox{1\width}{!}{
					$\sum_{X \in {\chi -\chi^*}} P(\xi_0 = X)d(\xi_0) $
				}
				\\ = &
				\resizebox{1\width}{!}{
					$\sum_{d=1}^n d \pi_0{(\chi_d)}$
				}
				\vspace{-0.7em}
				\label{equ:Ed0_mutation1}
			\end{aligned}
		\end{equation}	
		In the ENAS algorithm, the initial population follows a uniform distribution $\pi_0(\chi_d)={|\chi_d|}/({|\chi|-|\chi^*|})$. According to the drift analysis results and initial distance, i.e., \myref{con:ave_drift_t_mutation1} and \myref{equ:Ed0_mutation1},
		we can get the EHT of the ENAS algorithm by \myref{eq:et_theorem1}:
		\begin{equation}  \label{eq:et_theorem1}
			\begin{aligned}
				\mathbb{E}[T(\xi_0)] = & {\mathbb{E}[d(\xi_0)]}/{\bar{\Delta}_t} 
				\\ = &
				\frac{
					\resizebox{1\width}{!}{
						$\sum_{d=1}^n d \pi_0{(\chi_d)}$
					}
				}{
					\resizebox{1\width}{!}{  
						$ 
						\sum_{d=1}^n \Delta{(d)} {\pi_t{(\chi_d)}} /(1-\pi_t{(\chi^*)})$
					}
				}
				\\ \geq &
				\frac{
					n (1-\pi_t(\chi^*))
					\sum_{d=1}^n d \pi_0{(\chi_d)}
				}{
					\sum_{d=1}^n d
					\sum_{\gamma=1}^\lambda \gamma \pi_t(\chi_d^\gamma)
				}
			\end{aligned}
		\end{equation}
		
	\end{IEEEproof}

	\begin{theorem}
		For a ($\lambda$+$\lambda$)-ENAS algorithm using the one-bit mutation (\textbf{Mutation\#2}) and the non-repeated selection mechanism, its EHT is lower bounded by \myref{eq:lbEHT_M2}:
		\begin{equation}
			\frac{
				Q
				(1-\pi_t(\chi^*))
				\sum_{d=1}^n
				d
				\pi_0(\chi_d)
			}{
				\sum_{d=1}^n
				d
				\sum_{\gamma=1}^\lambda
				\gamma
				\pi_t(\chi_d^\gamma)
			}
			\label{eq:lbEHT_M2}
		\end{equation}
		where $n$ is the solution size, $Q=n_1+Ln_2$, $\lambda$ is the population size, $\pi_t{(\chi_d^\gamma)}$ and $\pi_t{(\chi_d)}$ can be calculated by population distribution $\pi_t$.
		\label{theorem:CEHTENAS_M2}
	\end{theorem}	
	
	The proof of the Theorem~\ref{theorem:CEHTENAS_M2} is similar to that of Theorem \ref{theorem:CEHTENAS_M1}. The difference is that the transition probability analysis needs to be solved in detail according to the mutation operator used in the algorithm. Due to space limitations, detailed proof will be provided in the supplementary materials.

	\subsection{($\lambda$+$\lambda$)-ENAS algorithm with multi-bit mutation}
	
	By using Lemmas~\ref{lemma:combination_q_tp_onebit_M3} and \ref{lemma:combination_q_tp_onebit_M4}, we show the lower bounds corresponding to the $q$-bit and bitwise mutation strategies we introduced in \uppercase\expandafter{\romannumeral3}-A. Compared with one-bit mutation, multi-bit mutation is more flexible in practical applications. Next, we give two theorems about EHT lower bound and their proof process of two ($\lambda$+$\lambda$)-ENAS algorithms using $q$-bit mutation and bitwise mutation, more precisely, that \textbf{Mutation\#3} and \textbf{Mutation\#4} are used respectively. 
	
	\begin{theorem}
		For a ($\lambda$+$\lambda$)-ENAS algorithm using the $q$-bit mutation (\textbf{Mutation\#3}) and the non-repeated selection mechanism, its EHT is lower bounded by \myref{eq:lbEHT_M3}:
		\begin{equation}
			\frac{
				\resizebox{1\width}{!}{
					$(1-\pi_t{(\chi^*)})
					\sum_{d=1}^n d \pi_0{(\chi_d)}$
				}
			}{
				\resizebox{0.9\width}{!}{  
					$
					\sum_{d=1}^n
					(
					q -
					\sum_{j=0}^{q-1} 
					(
					\mathop{\min}\limits_{d_x \in [d,n]} \{
					\sum_{d_y=d-j}^{d} 
					P_3^m(x,y)
					\}
					)^\lambda
					)
					\cdot
					\pi_t(\chi_d)
					$
				}
			}
			\label{eq:lbEHT_M3}
		\end{equation}
		where $n$ is the solution size, $\lambda$ is the population size, $q$ is the mutation bits, $P_3^m(x,y)$ represents the \textbf{Mutation\#3} transition probability of the parents $x$ to offspring $y$ with distance $d_y$ and can be derived from~\myref{eq:pmxy_M3}, and $\pi_t{(\chi_d)}$ is the population distribution $\pi_t$.
		
		\label{theorem:CEHTENAS_M3}
	\end{theorem}
	
	\begin{IEEEproof}
		Similar to the proof of Theorem \ref{theorem:CEHTENAS_M2}, the first step is to model the process of $(\lambda+\lambda)$-ENAS algorithm using $q$-bit mutation (\textbf{Mutation\#3}) as $\left\lbrace {\xi_t}\right\rbrace _{t=0}^{+\infty} ({\xi_t}\in \chi)$. The search space $\chi$ is also partitioned by the methods used in the above proof. Differently, the calculation processes of transition probability and drift caused by various mutation methods are different. Next, we will detail these calculation processes from four aspects: mutation transition probability, point-wise drift, average drift, and the EHT.
		
		1) At the $t$-th generation, $\xi_t$ can be any population $X$ in $\chi_k$.
		When $x$ executes the $q$-bit mutation, its \textbf{mutation transition probability} is calculated as follows. Once $X$ performs one step of mutation, the algorithm will generate a mutation population $Y$ belonging to $\chi_{k-q}\cup \chi_{k-q+1}\cup \cdots\cup\chi_{k+q}$. Regardless of how the variables are chosen, the distance of the new population $Y$ is between $[k-q, n]$. From~\myref{eq:pmxy_M3} we know that the distance of population $Y$ can be expressed as $d_Y=(k+q-j)$, where $j\in [0,2q]$. For any solution $x$ in $X$, we use $d_y=(k+q-j')$ to represent the distance of its offspring $y$, where there is $j'\leq j$. For a given value $j$, if $d_y$ is greater than or equal to value $d_Y$, i.e., $(k+q-j')\geq (k+q-j)$, then the probability that the distance value of offspring $y$ is greater than parent $x$ is $\sum_{j'\leq j}P^m(x,y)$, where $P^m(x,y)$ can be obtained by ~\myref{eq:pmxy_M3}. On the basis of that, if all the solutions have distance $d_y$ greater than or equal to $(k+q-j)$ and there exists at least one solution whose offspring distance equal to value $(k+q-j)$, there is $Y$ belonging to $\chi_{k+q-j}$ and the probability can be expressed by \myref{equ:E(P(Y|X))_M3}:
		\vspace{-0.3em}
		\begin{equation} 
			\label{equ:E(P(Y|X))_M3}
			\begin{aligned}
				& P(Y \in \chi_{k+q-j} \mid X \in \chi_k) 
				\\ = & 
				\prod_{x\in X} 
				\sum_{j'\leq j} P^m(x,y) - 
				\prod_{x\in X} 
				\sum_{j'\leq j-1} P^m(x,y)
			\end{aligned}
			\vspace{-0.1em}
		\end{equation}
		where $P^m(x,y)$ refers to the probability of an individual $x$ mutated to an individual $y$ with distance $(k+q-j')$ by using $q$-bit mutation, the left term of the minus is based on considering the distance $d_y$ of each solution in $X$ is greater than or equal to $(k+q-j)$, and the right term of the minus represents the probability that each $d_y$ is greater than $(k+q-j)$.
		
		2) Based on the above analysis, the \textbf{point-wise drift} from $\xi_t$ to $\xi_{t+1}$ can be calculated by \myref{con:drift_t_2_M3_without_d}:
		\vspace{-0.2em}
		\begin{equation}
			\begin{aligned}
				\Delta(X) 
				= & 
				\mathbb{E}[d(\xi_t)-d(\xi_{t+1}) \mid \xi_t = X]
				\\ = &  
				\resizebox{1\width}{!}{  
					$\sum_{Z \in \chi_{[(d_X-q):n]}}
					P(X,Z){(d_X-d_Z)} $
				}
				\\ \leq &
				\resizebox{1\width}{!}{  
					$\sum_{Z \in \chi_{[(d_X-q):(d_X-1)]}}P(X,Z){(d_X-d_Z)} $
				}
				\\ \leq &
				\resizebox{1\width}{!}{
					$\sum_{j=q+1}^{2q}
					P(Y \in \chi_{(d_X+q-j)}\mid X)\cdot (j-q)
					$
				}
			\end{aligned}
			\label{con:drift_t_2_M3_without_d}
			\vspace{-0.2em}
		\end{equation}
		where the second inequality is based on considering the most optimistic situation that population $Y$ is the same as $Z$. When population $X$ belongs to subspace $\chi_d$, the distance of $X$ is $d$, i.e., $d_X=d$. 
		By abbreviating $P(Y \in \chi_{(d+q-j)}\mid X \in \chi_d)$ of~\myref{equ:E(P(Y|X))_M3} as function $F(j)=f(j)-f(j-1)$ with respect to $j$, we can obtain \myref{equ:derivation}:
		\vspace{-0.2em}
		\begin{equation}
			\label{equ:derivation}
			\begin{aligned}
				&
				\sum_{j=q+1}^{2q} F(j)\cdot (j-q)
				=q\cdot f(2q) - \sum_{j=q}^{2q-1}f(j)
			\end{aligned}
			\vspace{-0.2em}
		\end{equation}
		Thus, $\Delta(X \in \chi_d)$ satisfies \myref{con:drift_t_2_M3}:
		\vspace{-0.5em}
		\begin{equation}
			\label{con:drift_t_2_M3}
			\begin{aligned}
				& \Delta(X \in \chi_d) 
				\leq
				\sum_{j=q+1}^{2q}
				P(Y \in \chi_{(d+q-j)}\mid X \in \chi_d)\cdot (j-q) 
				\\= & 
				q \prod_{x\in X} (\sum_{j'\leq 2q} P^m(x,y))
				-
				\sum_{j=q}^{2q-1}
				(
				\prod_{x\in X} (\sum_{j'\leq j} P^m(x,y))
				)
				\\ = &
				q \prod_{x\in X} (
				\sum_{d_y\geq d-q}P^m(x,y)
				)
				-
				\sum_{j=0}^{q-1} 
				(
				\prod_{x\in X}
				(\sum_{d_y\geq d-j} P^m(x,y))
				\\ \leq &
				q -
				\sum_{j=0}^{q-1} 
				(
				\mathop{\min}\limits_{d_x \in [d,n]} \{
				\sum_{d_y\geq d-j} P^m(x,y)\}
				)^\lambda
				, 
				\vspace{-0.5em}
			\end{aligned}
		\end{equation}
		where the first equation can be derived from~\myref{equ:E(P(Y|X))_M3} and~\myref{equ:derivation}, the left term of the second inequality can be obtained by considering $\sum_{d_y\geq d-q}P^m(x,y)\leq 1$ for any $x\in X$, and the right term of the second inequality is obtained by considering the minimum $P^m(\cdot,\cdot)$ for any solution in $X$. 
		It can be observed that ~\myref{con:drift_t_2_M3} is related to the distance function $d$. 
		Thus, for all of $X$ with the same distance, their point-wise drift values, i.e., $\Delta(X:d_X=d)$, have a general upper bound and can be uniformly denoted as $\Delta(d)$.
		
		3) Therefore, the \textbf{average drift} at the $t$-th generation can be calculated by \myref{con:ave_drift_t_M3}:
		\vspace{-0.3em}
		\begin{equation} 
			\begin{aligned}
				\bar {\Delta}_t 
				= & 
				\mathbb{E}[\mathbb{E}[d(\xi_t)-d(\xi_{t+1})\mid \xi_t=X\sim \pi_t]\mid X\in\chi-\chi^* ]
				\\ = &
				\sum_{d=1}^n (
				\frac{\pi_t{(\chi_d)}}{1-\pi_t{(\chi^*)}}
				\cdot
				\Delta{(d)}
				)
				\\ \leq &
				\sum_{d=1}^n 
				\frac{
					q -
					\sum_{j=0}^{q-1} 
					(
					\mathop{\min}\limits_{d_x \in [d,n]} \{
					\sum_{d_y\geq d-j} P^m(x,y)\}
					)^\lambda
				}{(1-\pi_t{(\chi^*)})/
					\pi_t{(\chi_d)}}
			\end{aligned}
			\label{con:ave_drift_t_M3}
			\vspace{-0.3em}
		\end{equation}

		4) Similar to the previous theorems, we assume that the initial population follows uniform distribution. Thus, the expectation of $d(\xi_0)$ can be derived from~\myref{equ:Ed0_mutation1}. According to the \textbf{Mutation\#3} transition probability~\myref{eq:pmxy_M3} and average drift~\myref{con:ave_drift_t_M3}, the EHT satisfies \myref{eq:et_theorem2}:
		\begin{equation} \label{eq:et_theorem2}
			\begin{aligned}
				& \mathbb{E}[T(\xi_0)] = {\mathbb{E}[d(\xi_0)]}/{\bar{\Delta}_t} 
				\\ \geq &
				\frac{
					\resizebox{0.9\width}{!}{
						$(1-\pi_t{(\chi^*)})
						\sum_{d=1}^n d \pi_0{(\chi_d)}$
					}
				}{
					\resizebox{0.9\width}{!}{  
						$
						\sum_{d=1}^n
						(
						q -
						\sum_{j=0}^{q-1} 
						(
						\mathop{\min}\limits_{d_x \in [d,n]} \{
						\sum_{d_y\geq d-j}
						P_3^m(x,y)
						\}
						)^\lambda
						)
						\cdot
						\pi_t(\chi_d)
						$
					}
				}
			\end{aligned}
		\end{equation}
	\end{IEEEproof}

	\begin{theorem}
		For a ($\lambda$+$\lambda$)-ENAS algorithm using the bitwise mutation and the non-repeated selection mechanism, its EHT is lower bounded by \myref{eq:lbEHT_M5}:
		\begin{equation}
			\frac{
				\resizebox{0.9\width}{!}{
					$(1-\pi_t{(\chi^*)})
					\sum_{d=1}^n d \pi_0{(\chi_d)}$
				}
			}{
				\resizebox{0.9\width}{!}{  
					$
					\sum_{d=1}^{n}
					(
					d-
					\sum_{D=1}^{d}
					(
					\mathop{\min}\limits_{d_x \in [d,n]}
					\{
					\sum_{d_y\geq D}
					P_4^m(x,y)
					\}
					)^\lambda
					)
					\cdot
					\pi_t{(\chi_d)}
					$
				}
			}
			\label{eq:lbEHT_M5}
		\end{equation}
		where $n$ is the solution size, $\lambda$ is the population size, $P_4^m(x,y)$ represents the \textbf{Mutation\#4} transition probability of the parents $x$ to offspring $y$ and can be derived from~\myref{eq:pmxy_M5}, and $\pi_t{(\chi_d)}$ is the population distribution $\pi_t$.
		\label{theorem:CEHTENAS_M5}
		
	\end{theorem}
	
	For each solution, bitwise mutation (\textbf{Mutation\#4}) implies that the individual independently modifies each bit with probability $p=1/n$ to produce a new solution. Since each bit has the same probability of being mutated, which follows the same principle as \textbf{Mutation\#3}, we can further get the lower bound of EHT based on Theorem~\ref{theorem:CEHTENAS_M5}. Due to space limitations, detailed proof will be provided in the supplementary materials.

    \section{Case Study}

        \begin{figure}[!t]
		\centering
		\subfigure[]{
			\includegraphics[width=0.43\linewidth]{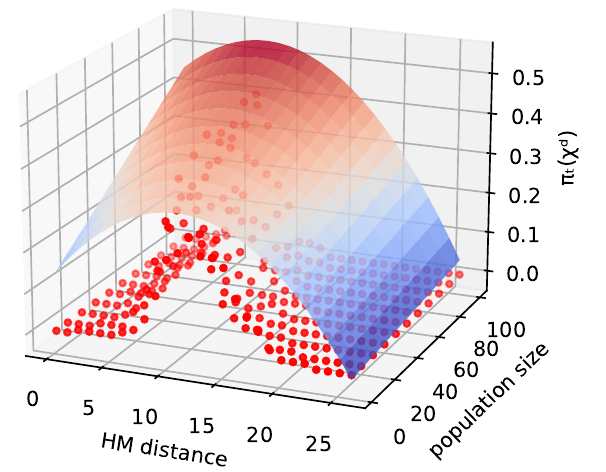}
			\label{mutation4_fitting}
		}
		\subfigure[]{
			\includegraphics[width=0.43\linewidth]{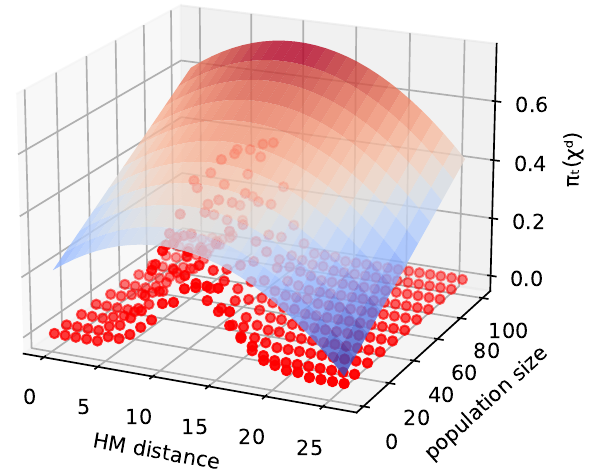}
			\label{mutation5_fitting}
		}
		\subfigure[]{
			\includegraphics[width=0.43\linewidth]{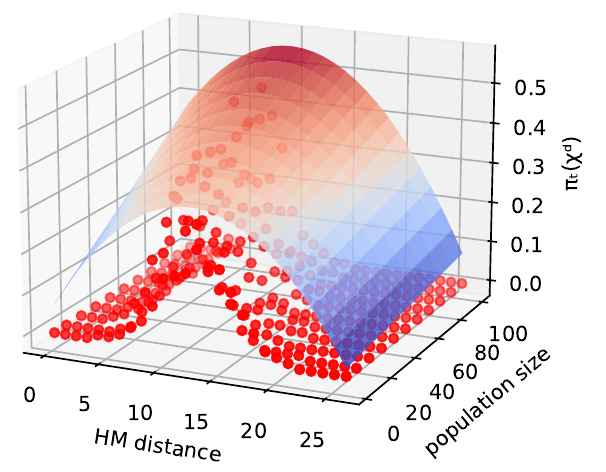}
			\label{mutation6_fitting}
		}
		\subfigure[]{
			\includegraphics[width=0.43\linewidth]{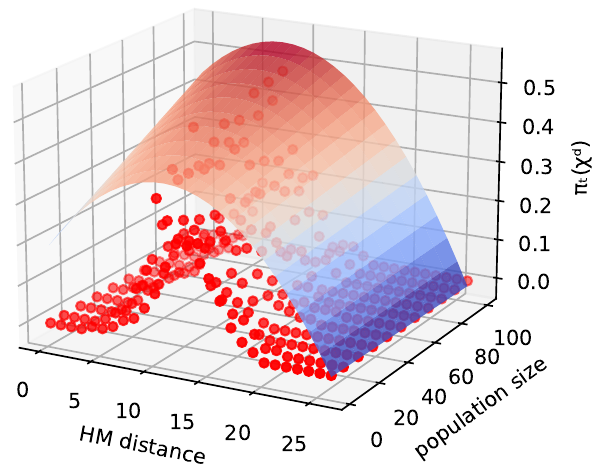}
			\label{mutation7_fitting}
		} 
		\caption{
			The visualization results of the sampling data (red dots) and fitting surfaces (colored surfaces) of $\pi_t$ for different ENAS algorithms, where the HM distance represents the Hamming distance of the population.
			(a) The fitting result (corresponding to $Z_1$) of $\pi_t$ for ENAS algorithm with \textbf{Mutation\#1}.
			(b) The fitting result (corresponding to $Z_2$) of $\pi_t$ for ENAS algorithm with \textbf{Mutation\#2}.
			(c) The fitting result (corresponding to $Z_3$) of $\pi_t$ for ENAS algorithm with \textbf{Mutation\#3}(q=2).
			(d) The fitting result (corresponding to $Z_4$) of $\pi_t$ for ENAS algorithm with \textbf{Mutation\#4}.
		}
		\label{fig:fittings}
	\end{figure}
	
	\begin{figure}[!t]
		\centering
		\subfigure[]{
			\includegraphics[width=0.46\linewidth]{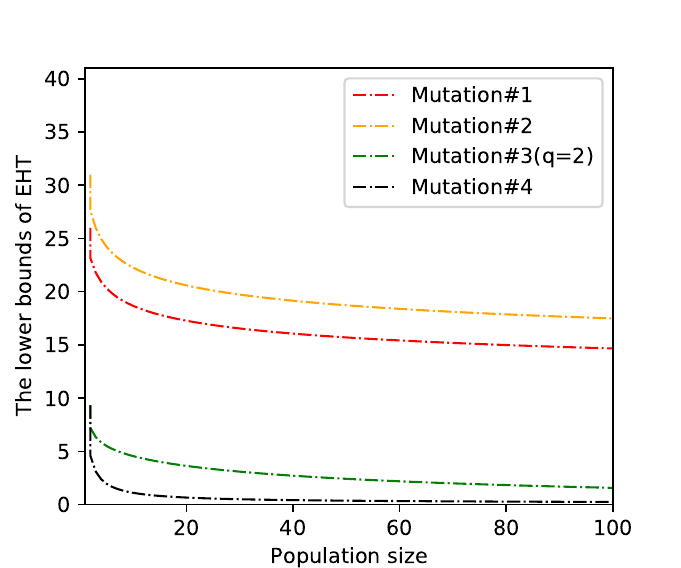}
			\label{eht}
		}
		\subfigure[]{
			\includegraphics[width=0.46\linewidth]{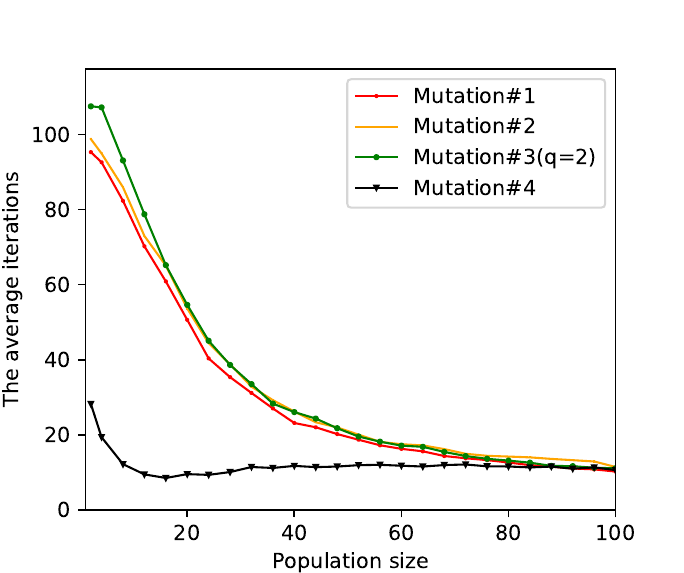}
			\label{real}
		}
		\caption{
			The theoretical and experimental running times (iterations) of the ($\lambda$+$\lambda$)-ENAS algorithms using mutation operators \textbf{Mutation\#1}, \textbf{Mutation\#2}, \textbf{Mutation\#3}, and \textbf{Mutation\#4}, respectively, where the parameters of search space are set to $n=26$ and $L=2$.
			(a) The lower bounds of EHT of ($\lambda$+$\lambda$)-ENAS algorithms with population sizes $\lambda$ ranging from 1 to 100.
			(b) The average iterations count of ($\lambda$+$\lambda$)-ENAS algorithms with population sizes $\lambda$ ranging from 1 to 100 with step size four.
		}
		\label{fig:theoretical_experimental_real_eht}
	\end{figure}

	In this section, we conduct case studies on the widely used NAS-Bench-101 benchmark~\cite{ying2019bench} to validate the validity of the proposed running time analysis method. NAS-Bench-101 is the first architecture dataset for NAS and provides a cell-based search space. It consists of a mapping table that contains 432k unique CNN architectures. Each architecture in NAS-Bench-101 has a maximum of seven nodes (i.e., $v\leq 7$) and a maximum of three kinds of operations (i.e., $L=2$). The NAS-Bench-101 dataset includes training results for each architecture on the CIFAR-10 dataset, such as validation accuracy, test accuracy, and training time. The dataset is designed for NAS research to alleviate the need for intensive computing resources. We use this dataset to perform theoretical research on the EHT of ENAS algorithms. The NAS problem can be viewed as searching for the best architecture from the search space in NAS-Bench-101, and thus can be called the NAS-Bench-101 architecture searching problem. In the following, we will present the EHT lower bounds and empirical running times for four ($\lambda$+$\lambda$)-ENAS algorithms (with the different mutation operators in \textbf{Mutation\#1-Mutation\#4}).  
	
	First, we arrange the four EHT results for observation. According to the Theorems \ref{theorem:CEHTENAS_M1} to 
	\ref{theorem:CEHTENAS_M5}, the EHT lower bounds of the four ($\lambda$+$\lambda$)-ENAS algorithms using different mutation strategies can be further derived \myref{eq:exp}: 
	\vspace{-0.5em}
	\begin{equation} \label{eq:exp}
		\rm{EHT} \geq
		\begin{aligned}
			\begin{cases}
				\frac{
					n
				}{
					\sum_{d=1}^n
					d
					\sum_{\gamma=1}^\lambda
					\gamma R_1(d,\gamma)
				}
				\\
				\frac{
					n+v-2
				}{
					\sum_{d=1}^n
					d
					\sum_{\gamma=1}^\lambda
					\gamma R_1(d,\gamma)
				}
				\\
				\frac{
					1
				}{
					\resizebox{0.8\width}{!}{  
						$
						\sum_{d=1}^n
						(
						q-
						\sum_{j=0}^{q-1}
						(
						\mathop{\min}\limits_{d_x \in [d,n]} \{
						\sum_{j'\geq d-j}
						P_3^m(x,y)
						\}
						)^\lambda
						)
						\cdot
						R_2(d)
						$
					}
				}
				\\
				\frac{
					1
				}{
					\resizebox{0.82\width}{!}{  
						$
						\sum_{d=1}^{n}
						(
						d-
						\sum_{D=1}^{d}
						(
						\mathop{\min}\limits_{d_x \in [d,n]}
						\{
						\sum_{d_y\geq D}
						P_4^m(x,y)
						\}
						)^\lambda
						)
						\cdot
						R_2(d)
						$
					}
				}
			\end{cases}
		\end{aligned}
	\end{equation}
	where $R_1(d,\gamma)$ is shorthand for the common expression 
	$\pi_t(\chi_d) (\gamma|\chi_d^\gamma|/|\chi_d|) / [(|\chi|-|\chi^*|)\sum_{k=1}^n k |\chi_k|/|\chi|^2]$
	, $R_2(d)$ is shorthand for the common expression 
	$\pi_t(\chi_d) / [(|\chi|-|\chi^*|)\sum_{k=1}^n k |\chi_k|/|\chi|^2]$
	, $P_3^m(x,y)$ and $P_4^m(x,y)$ can be derived from~\myref{eq:pmxy_M3} and~\myref{eq:pmxy_M5}. For calculating the EHT lower bounds, we execute the surface fitting experiments to fill in the missing $\pi_t$ in the EHT results.

	Second, according to the proposed theorems, it can be inferred that EHT is negatively correlated with the population distribution $\pi_t(\chi_d)$. To obtain each lower bound of EHT, we execute statistical experiments to derive the upper bound of $\pi_t(\chi_d)$ for each ENAS algorithm. In the experiments, the parameters about the problem size are set to $v=7$ (i.e., $n=26$) and $L=3$. This is because the architectures with $v=7$ account for 84.76$\%$ of all architecture in the NAS-Bench-101 dataset, which takes the majority among the overall architectures. Thus, we choose the architectures with $v=7$ from NAS-Bench-101 in the experiments. The operator settings of the ENAS algorithm are based on the proposed theorems, i.e., one-bit mutation operators (\textbf{Mutation\#1} and \textbf{Mutation\#2}), $q$-bit mutation operator (\textbf{Mutation\#3}) where $q=2$, and bitwise mutation operator (\textbf{Mutation\#4}). The population size $\lambda$ is selected from $\{2,10,20,30,40,50,60,70,80,90,100\}$. For each value of $\lambda$, we conduct experiments (detailed in~\ref{Subsection:fitting}) to collect sampling data, where each data represents the upper bound probability of the population's Hamming distance being equal to a specific.
		
	After collecting the sampling data, we utilize the non-linear programming solver (SLSQP) provided by Python to solve the surface fitting problem. The fitting results can be represented by \myref{eq:last}: 
	\begin{equation} \label{eq:last}
		\left\{
		\begin{aligned}
			& \resizebox{0.95\width}{!}{${Z_1} = -0.0015x^2 - 0.0001xy + 0.0333x + 0.0031y + 0.1357$} \\
			& \resizebox{0.95\width}{!}{${Z_2} = -0.0013x^2 + 0.0266x + 0.004y + 0.2026$}	\\
			& \resizebox{0.92\width}{!}{${Z_3} = -0.0018x^2 - 0.0001xy + 0.0451x + 0.0032y + 0.0619$}	\\
			& \resizebox{0.92\width}{!}{${Z_4} = -0.0017x^2 - 0.0001xy + 0.0368x + 0.0025y + 0.1888$}	\\
		\end{aligned}
		\right.
	\end{equation}
	where each $Z$ in $\{Z_1,Z_2,Z_3,Z_4\}$ denotes the upper bound of $\pi_t(\chi_d)$ for four different ENAS algorithms, $x$ represents the Hamming distance of population, and $y$ represent the population size. These equations illustrate the relationship among the population size, population distance values, and the lower bound of the population distribution. Fig~\ref{fig:fittings} depicts the visualization results of the sampled data and fitting surfaces, where the red dots represent the sampled data and the colored surface represents the fitting surface. Subsequently, we can substitute the fitting results as the population distribution into the lower bound formula of EHT. These lower bounds are then presented using the visualization method shown in Fig.~\ref{eht}. Furthermore, we executed 1000 runs of each configured ($\lambda$+$\lambda$)-ENAS algorithm to obtain the experimental running time (i.e., the average iterations). Fig.~\ref{real} depicts the average iterations for each algorithm with different population sizes. By comparing the EHT results with experimental running time results, we observe that the lower bounds of EHT are consistently lower than the corresponding experimental running time. This demonstrates the validity of the proposed method for estimating the EHT of the ENAS algorithms.
        
	Moreover, both theoretical analysis and empirical experiments have proven that the superposition of \textbf{Mutation\#4} and population brings benefit to the running time. This is because the bitwise operator causes each bit of the solution to have a probability of changing, thereby expanding the range of searched offspring in each generation. As the population size grows, the range of offspring also increases, which in turn increases the probability of the algorithm escaping from local optima. However, the combination of \textbf{Mutation\#4} and population also has its limitation, i.e., a significant drop only around population sizes close to 10 and 15, as observed from both Figs.~\ref{eht} and~\ref{real}.

	\section{Conclusion}
	
	The paper aims to investigate the method of integrating theory and experiment to estimate the expected running time of the ($\lambda$+$\lambda$)-ENAS algorithm with various operators, so that the minimum number of generations for algorithms can be appropriately scheduled in advance. The investigation has been achieved by estimating the lower bound of the EHT through the proposed CEHT-ENAS method. Specifically, in the proposed CEHT-ENAS, we present the common configuration to encode the complex structure of deep architectures and then introduce the partition method to measure the progress. Following, we give five lemmas to estimate the transition probabilities of different mutation operators under binary encoding and combination encoding methods. 
	Furthermore, the paper provides the mathematical analysis of the lower bounds of EHT regarding ($\lambda$+$\lambda$)-ENAS using different mutation operators. Specifically, two theorems are given from the perspective of one-bit mutation, i.e., Theorems~\ref{theorem:CEHTENAS_M1}-\ref{theorem:CEHTENAS_M2}, and two theorems are given from the perspective of multi-bit mutation (including bitwise mutation), i.e., Theorems~\ref{theorem:CEHTENAS_M3}-\ref{theorem:CEHTENAS_M5}. In addition, we utilize the surface fitting method to obtain the bounds of the population distribution, and the bounds of EHT were further estimated by using the proposed theorems. The case study confirms that the lower bound on EHT of ENAS can be effectively estimated by the proposed method of integrating theory and experiment. The results presented here can be viewed as an initial exploration of analyzing the running time of the ENAS algorithm with mutation.
 
    In this work, we assume that the optimal solution of the NAS problem is unique. In practice, most NAS algorithms, including NAS benchmarks, have a unique optimal solution. However, there are also some NAS algorithms which may have multiple optimal solutions, such as the multi-objective NAS algorithms. To this end, the proposed method will not be applicable. In the future, we will invest efforts to extend the applicable scenarios of the proposed method. Furthermore, the crossover operators are also important in solving NAS problems. Further studies will also be performed to analyze the ENAS algorithm with crossover operator.

	\vspace{-0.5em}




	\bibliographystyle{IEEEtran}
	\bibliography{reference}
	
	\vspace{-5em}
	\begin{IEEEbiography}[             {\includegraphics[width=1in,height=1.25in,clip,keepaspectratio]{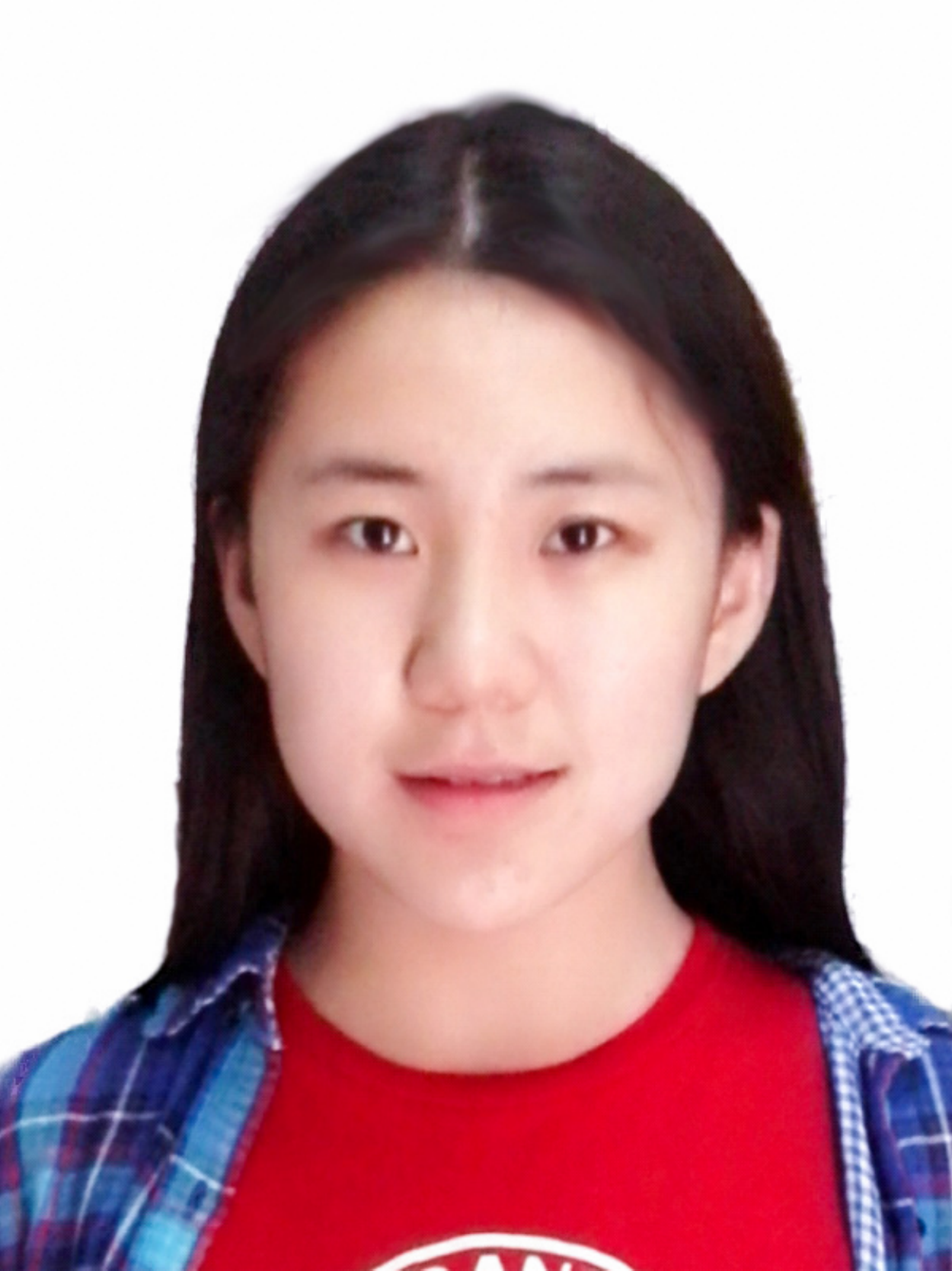}}]{Zeqiong Lv} 
		received her M.E. degree in Computer Science from Xihua University, Chengdu, China, in 2021. She is currently pursuing a Ph.D. degree in Computer Science from Sichuan University, Chengdu, China. Her current research interests include evolutionary computation, neural networks, and theoretical analysis of evolutionary algorithms.
	\end{IEEEbiography}
	\vspace{-3em}
	\begin{IEEEbiography}[             {\includegraphics[width=1in,height=1.25in,clip,keepaspectratio]{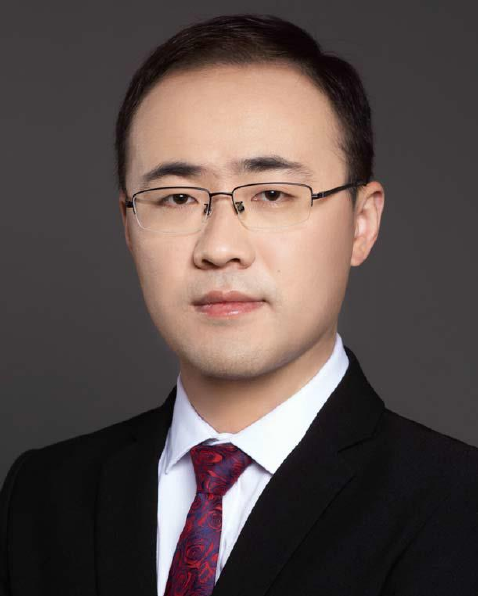}}]{Chao Qian} 
		(Senior Member, IEEE) is an Associate Professor in the School of Artificial Intelligence, Nanjing University, China. He received the BSc and PhD degrees in the Department of Computer Science and Technology from Nanjing University. After finishing his PhD in 2015, he became an Associate Researcher in the School of Computer Science and Technology, University of Science and Technology of China, until 2019, when he returned to Nanjing University.
		
		His research interests are mainly theoretical analysis of evolutionary algorithms (EAs), design of safe and efficient EAs, evolutionary learning, and application of EAs to solve real-world complex problems. He has published one book ``Evolutionary Learning: Advances in Theories and Algorithms", and over 50 papers in top-tier journals (AIJ, ECJ, TEvC, Algorithmica, TCS) and conferences (AAAI, IJCAI, NeurIPS, ICLR). He has won the ACM GECCO 2011 Best Theory Paper Award, the IDEAL 2016 Best Paper Award, and the IEEE CEC 2021 Best Student Paper Award Nomination. He is an associate editor of IEEE Transactions on Evolutionary Computation, and was the chair of IEEE CIS Task Force on Theoretical Foundations of Bio-inspired Computation. He has been invited to give an Early Career Spotlight Talk at IJCAI 2022, and will be a Program Co-Chair of PRICAI 2025. He is a recipient of the National Science Foundation for Excellent Young Scholars (2020), and CCF-IEEE CS Young Computer Scientist Award (2023).
	\end{IEEEbiography}
	\vspace{-3em}
	\begin{IEEEbiography}[{\includegraphics[width=1in,height=1.25in,clip,keepaspectratio]{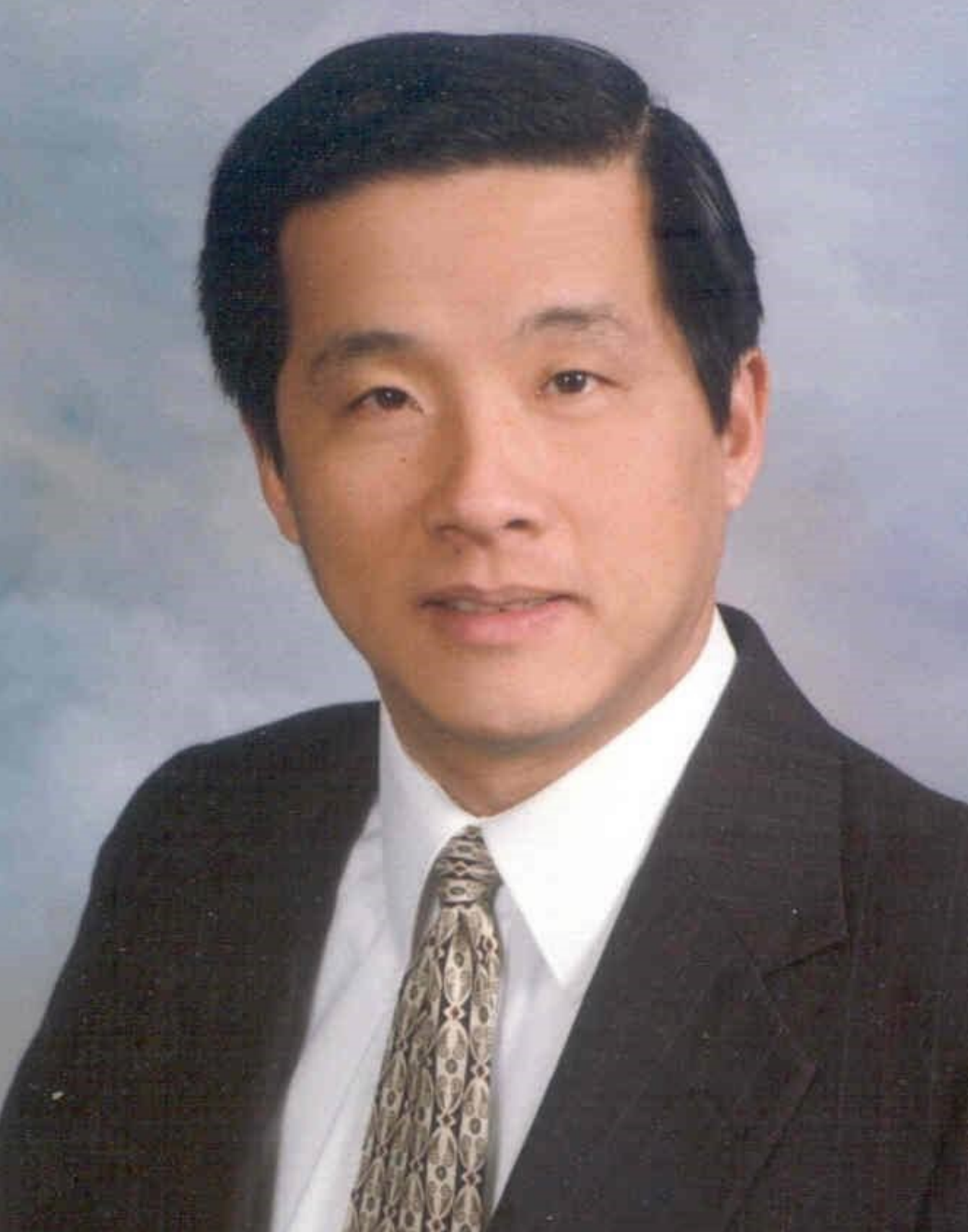}}]{Gary G. Yen}
		(Fellow, IEEE) received a Ph.D. degree in electrical and computer engineering from the University of Notre Dame in 1992. Currently he is a Regents Professor in the School of Electrical and Computer Engineering, Oklahoma State University (OSU). Before joined OSU in 1997, he was with the Structure Control Division, U.S. Air Force Research Laboratory in Albuquerque. His research interest includes intelligent control, computational intelligence, conditional health monitoring, signal processing and their industrial/defense applications.
		
		Dr. Yen was an associate editor of the \textit{IEEE Control Systems Magazine, IEEE Transactions on Control Systems Technology}, \textit{Automatica}, \textit{Mechantronics}, \textit{IEEE Transactions on Systems, Man and Cybernetics, Parts A and B} and I\textit{EEE Transactions on Neural Networks}. He is currently serving as an associate editor for the \textit{IEEE Transactions on Evolutionary Computation} and the \textit{IEEE Transactions on Cybernetics}. He served as the General Chair for the \textit{2003 IEEE International Symposium on Intelligent Control} held in Houston, TX and \textit{2006 IEEE World Congress on Computational Intelligence} held in Vancouver, Canada. Dr. Yen served as Vice President for the Technical Activities in 2005-2006 and then President in 2010-2011 of the IEEE Computational intelligence Society. He was the founding editor-in-chief of the \textit{IEEE Computational Intelligence Magazine}, 2006-2009. In 2011, he received Andrew P Sage Best Transactions Paper award from \textit{IEEE Systems, Man and Cybernetics Society} and in 2014, he received Meritorious Service award from \textit{IEEE Computational Intelligence Society}.
	\end{IEEEbiography}
	\vspace{-3em}
	\begin{IEEEbiography}[             {\includegraphics[width=1in,height=1.25in,clip,keepaspectratio]{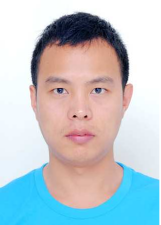}}]{Yanan Sun} 
		(Senior Member, IEEE) received his Ph.D. degree in Computer Science from Sichuan University, Chengdu, China in 2017. He is currently a Professor with the College of Computer Science, Sichuan University, Chengdu, China. His current research interests include evolutionary computation, neural networks, and their applications on neural architecture search. He designed the indicator of ``GPU Day", which has been widely used among the community of neural architecture search. He was ranked as ``World's Top 2\% Scientists 2021"  collectively released by Stanford University and Springer. He is associate editors of IEEE Transactions on Evolutionary Computation and IEEE Transactions on Neural Networks and Learning Systems.
	\end{IEEEbiography}

	\vfill


	
\end{document}


\title{Analyzing the Expected Hitting Time of Evolutionary Computation-based Neural Architecture Search Algorithms 
		
		(Supplementary Material)}
	
	\author{Zeqiong Lv,
		Chao Qian,~\IEEEmembership{Senior Member,~IEEE,}
		Gary~G.~Yen,~\IEEEmembership{Fellow,~IEEE,} 
		and Yanan Sun,~\IEEEmembership{Senior Member,~IEEE}
		\thanks{Zeqiong Lv and Yanan Sun are with the College of Computer Science, Sichuan University, Chengdu 610065, China (e-mails: zq\_lv@stu.scu.edu.cn; ysun@scu.edu.cn).}
		\thanks{Chao Qian is with the State Key Laboratory for Novel Software Technology, Nanjing University, Nanjing 210023, China (e-mail: qianc@nju.edu.cn).}
		\thanks{Gary G. Yen is with the School of Electrical and Computer Engineering, Oklahoma State University, Stillwater, OK 74078 USA (e-mail:gyen@okstate.edu).}
	}
	
	\maketitle
	
	\section{Introduction}
	This part aims to provide detailed proofs of lemmas and theorems, which are omitted in our original paper due to space limitations. Specifically, Section II presents the proofs of four lemmas concerning transition probability; Section III details the proofs of two theorems regarding EHT lower bounds. 
	
	\section{Detailed Proofs of Lemmas}
	
	This part will provide detailed proofs for Lemmas 2-5.

	\begin{IEEEproof}[Proof of Lemma 2]
        We first analyze the case of a binary encoding EA with one-bit mutation. When an individual $x$ executes the one-bit mutation, the Hamming distance of the new individual $y$ differs from $x$ by one. Since each solution randomly flips one bit, which can either be the same as or different from the optimal individual, the distance of $x$ may either decrease or increase by one. The probability of this change depends on the size of the individual. For example, consider the optimal solution represented by the string $(11001)$. The solution $(10101)$, which has a distance value of two, can generate one new string from the population $\left\lbrace (00101), (11101), (10001), (10111), (10100)\right\rbrace $ with a transition probability of $1/5$. The distance of the new string can be one with a probability of $2/5$ or three with a probability of $3/5$. As a result, we can view the transition probability of the distance between two neighboring generations ($d_x$ to $d_y$) as the mutation transition probability of $x$ to $y$, i.e., $P^m(x,y)=P(d_x,d_y)={{d_x\choose 1-i}{{n-d_x}\choose i} /{n \choose 1}}$, where $d_x$ and $d_y \in \left\lbrace {d_x-1+2i\mid i=0,1}\right\rbrace $ represent the distance of $x$ and $y$, respectively, and $n$ is the individual length.
		
		Based on the analysis, the mutation transition probability of individual $x$ using $q$-bit mutation can be generalized as:
		\begin{equation*} 
			\label{eq:pmxy1}
			\resizebox{1\width}{!}{$P^m(x,y) = P(d_x,d_y) = {{d_x \choose {q-i}} {{n-d_x}\choose {i}}}/{{n \choose q}}$}
		\end{equation*}
		where $d_y \in \left\lbrace d_x-q+2i\mid i=0,1,2, \ldots, q\right\rbrace$.

        According to the combination encoding method, $x$ can be divided into two parts: $x_1={\left\lbrace 0,1\right\rbrace }^{n_1}$ and $x_2={\left\lbrace 0,1,...,L\right\rbrace }^{n_2}$, with respective distance values denoted as $d_1$ and $d_2$. Based on this division, we continue to prove.

		For individual $x$ performing \textbf{Mutation\#1}, the probability of randomly selecting each bit is the same, i.e., $1/n$. There are two cases for randomly flipping a bit: the probability of randomly selecting one bit belonging to the first $n_1$ part is $n_1/n$, and the probability of randomly selecting one bit belonging to the last $n_2$ is $n_2/n$. Each case can be further analyzed in detail.
		For the first case, if the chosen bit is from the mismatched bits (totally $d_1$ bits), then after the individual $x$ mutates to $y$, the distance reduces by one; otherwise, the chosen bit is from the correct bits (totally $n_1-d_1$ bits), and then, the distance increases by one. 
		For the second case, if the chosen bit is from the mismatched bits (totally $d_2$ bits), then the bit may mutate into one of the $(L-1)$ mismatches, and the distance of offspring still is $d$, or the bit may mutate into the correct bit and the distance reduces by one; otherwise, the chosen bit is from the correct bits (totally $n_2-d_2$ bits), and the bit mutates into one of the $L$ mismatches, then the distance increases by one. Therefore, considering all the above, $P^m(x,y)$ of \textbf{Mutation\#1} can be formalized by \myref{eq:pmxy2_1}.
		\begin{equation} \label{eq:pmxy2_1}
			P_1^m(x, y) = 
			\begin{cases}
				(d_1L+d_2)/{(nL)}, & {d_y=d_x-1} \\ 
				{d_2 (L-1)}/{(nL)}, &{d_y=d_x} \\
				1 - d/n, & d_y=d_x+1.
			\end{cases}
			\vspace{-1.5em}
		\end{equation}
	\end{IEEEproof}

	\begin{IEEEproof}[Proof of Lemma 3]
				When $x$ performing \textbf{Mutation\#2}, there will be $Q=n_1+Ln_2$ possibilities for the generated offspring $y$. If the mutated bit belongs to the first $n_1$ part, the distance $d_y$ can be either $d_x+1$ or $d_x-1$. In the former case, there are $n_1-d_1$ possibilities, while in the latter case, there are $d_1$ possibilities. On the other hand, if the mutated bit belongs to the last $n_2$ part, the distance $d_y$ can be $d_x+1$, $d_x$, or $d_x-1$. The number of possibilities for each case is $(n_2-d_2)L$, $d_2(L-1)$, and $d_2$, respectively. Therefore, the mutation transition probability $P^m(x,y)$ for \textbf{Mutation\#2} can be formulized by \myref{eq:pmxy2}.
				\begin{equation} \label{eq:pmxy2}
					P_2^m(x, y) = 
					\begin{cases}
						{d_x}/{Q}, & {d_y=d_x-1} \\ 
						{d_2 (L-1)}/{Q}, &{d_y=d_x} \\
						1 - ({d_1 + L d_2})/{Q}, & d_y=d_x+1.
					\end{cases}
					\vspace{-1em}
				\end{equation}	
	\end{IEEEproof}
	
	\begin{IEEEproof}[Proof of Lemma 4]
		For the convenience of description, we call the bits selected for mutation as mutation bits. 
		At the same time, we sort each bit in the individual $x$ according to whether it matches the optimal solution correctly, so we get the following sequence of individual bits: the first $d_1$ bits are the mismatched bits of $x$, the $(d_1+1)$ to $n_1$ bits are correctly matched bits, the $(n_1+1)$ to $(n_1+d_2)$ bits are mismatched bits, and the $(n_1+d_2+ 1)$ to $n$ bits are correctly matched bits.
		
		Performing the $q$-bit mutation (\textbf{Mutation\#3}), there will be ${n\choose q}=\sum_{z=0}^{\min\{q,n_2\}} {n_1\choose {q-z}} {n_2\choose z}$ possibilities of mutation bits selection. For each possibility, we separately consider the selected mutation bits for each part (i.e., the first $n_1$ part and the last $n_2$ part, respectively) according to the combination encoding of the individual. 
		The detailed calculation of the possibilities of each part is shown in \myref{eq:Cnq}: 
		\vspace{-0.5em} 
		\begin{equation} \label{eq:Cnq}
			\begin{aligned}
				\begin{cases}
					{n_1\choose {q-z}}
					=
					\sum_{a=0}^{\min\{q-z,d_1\}}
					{d_1\choose {\underbrace{a}_{d-1}}}
					{{n_1-d_1}\choose {\underbrace{q-z-a}_{d+1}}}
					\\
					{n_2\choose z}
					=
					\sum_{b=0}^{\min\{z,d_2\}}
					{d_2\choose \underbrace{b}_{d-1 \ \rm{or}\ d}}
					{{n_2-d_2}\choose \underbrace{z-b}_{d+1}}
				\end{cases}
			\end{aligned}
			\vspace{-0.5em} 
		\end{equation}
		Following that, we thoroughly explain the parameters involved in the formula, the distance value after mutation, and the mutation transition probability.
		First, the parameters in the proceeding formula are associated with the process of selecting $q$ mutation bits. Depending on whether each bit is in the first $n_1$ bits or the last $n_2$ bits, we give the following instructions: 1) there are $z$ bits that belong to the last $n_2$ bits; we assume that the number of ``mismatched bits" is $b\in [0,\min\{z,d_2\}]$, and the number of ``correctly matched bits" is $z-b$; 2) furthermore, for the above $b$ bits, $c$ bits will be successfully mutated, that is, mutated into matched bits by probability $p(c)=\binom{b}{c}(1/L)^{c}(1-1/L)^{b-c}$;
		3) there are $q-z$ bits belonging to the first $n_1$ bits, and further, we assume that the number of ``mismatched bits" is $a\in [0,\min\{q-z,d_1\}]$, then the number of bits that belongs to the correct matching bits is $q-z-a$. 
		
		For each bit in $q$ bits, if it is selected from $n-d$ correctly matched bits, the bit must become a mismatched bit after mutation, and the corresponding distance is increased by one; if the bit is selected from the $d_1$ mismatched bits, it must be mutated successfully, and the corresponding distance is reduced by one; if the bit is selected from the $d_2$ mismatched bits, it has a $(1/L)$ probability of mutating successfully and a $(1-1/L)$ probability of mutating into the wrong bit, corresponding to that the distance reduces by one or keeps unchanged.
		That is to say, after an individual with distance $d$ performs the mutation, the distance of its offspring is $d_y=d_x+q-(2a+b+c)$, and the $(2a+b+c)$ is abbreviated as $j\in[0,2q]$. 
		Thus, we can get the transition probability $P_3^m(x,y)$ that \textbf{Mutation\#3} makes offspring with distance $d_y=d_x+q-j$, as shown in~\myref{eq:pmxy_M3}:
		\begin{equation}\label{eq:pmxy_M3}
			\begin{cases}
				P_3^m(x,y) = 
				\sum_{z=0}^{\min{\{q,n_2\}}}
				\sum_{a=0}^{\min{\{q-z,d_1\}}}
				\sum_{b=0}^{\min{\{z,d_2\}}}
				\frac{
					W
				}{
					\binom{n}{q}
				}
				\\
				W = 
				\binom{d_1}{a}
				\binom{n_1-d_1}{q-z-a}
				\binom{d_2}{b}
				\binom{n_2-d_2}{z-b}
				\binom{b}{c}
				\left(\frac{1}{L}\right)^c
				\left(1-\frac{1}{L}\right)^{b-c}
				\\
				c=j-2a-b
			\end{cases}
		\end{equation}
		Note that since some combinations of $a$ and $b$ will not satisfy $j$, i.e., $c< 0$ or $c> b$, so at this point there is $\binom{b}{c}=0$.
	\end{IEEEproof}

	\begin{IEEEproof}[Proof of Lemma 5]
		After individual $x$ performs the bitwise mutation (\textbf{Mutation\#4}), the number of its mutation bits is between $[0,n]$, and the probability of any mutation bits $q$ is $\binom{n}{q} p_m^q (1-p_m)^{n-q}$, where $p_m=1/n$ is the mutation rate. Since each bit mutates with the same probability $p_m$, the mutation of individual $x$ follows the ``bit-based fair mutation" principle. Concurrently, the probability of mutating to offspring $y$ is the same as \textbf{Mutation\#3} under the assumption that the mutation bits is $q$. Thus, the bitwise mutation transition probability $P_4^m(x,y)$ can be calculated by \myref{eq:pmxy_M5}:
		\begin{equation} \label{eq:pmxy_M5}
			\begin{aligned}
				P_4^m(x,y) = & 
				\sum_{q=0}^{n}
				\binom{n}{q} 
				\left(\frac{1}{n}\right)^q \left(1-\frac{1}{n}\right)^{n-q}
				P_3^m(x,y)
			\end{aligned}
			\vspace{-0.5em}
		\end{equation} 
		where $P_3^m(x,y)$ is derived by~\myref{eq:pmxy_M3}.
    \end{IEEEproof}

	\section{Detailed Proofs of Theorems}
	
	This part will provide detailed proofs for Theorems 2 and 4.

	\begin{IEEEproof} [Proof of Theorem 2]
		The transition probability analysis and drift calculation process need to be solved in detail according to the mutation operator used in the algorithm. And the specific steps are as follows:
		
		1) At the $t$-th generation, the \textbf{transition probability} calculation result shown in \myref{equ:E(P(Y|X))_M2},
		\begin{equation} 
			\begin{aligned}
				& P(Y \in \chi_{k-1} \mid X \in \chi_k) 
				\\ = & \mathbb{E} [P(Y \in \chi_{k-1} \mid X \in \chi_k^\gamma)]
				= 
				\mathbb{E}[1-(1-P^m(k,k-1))^\gamma] 
				\\ = &
				\resizebox{1\width}{!}{
					${
						{\sum_{\gamma = 1}^{\lambda}(1-(1-{k}/{Q})^\gamma)}
						P(X\in\chi_k^\gamma) / P(X\in\chi_k) 
					}$ 
				}
				\\ \leq &
				\frac{k}{Q}
				\sum_{\gamma=1}^\lambda
				\gamma
				\frac{\pi_t(\chi_k^\gamma)}{\pi_t(\chi_k)}
			\end{aligned}
			\label{equ:E(P(Y|X))_M2}
		\end{equation}
		where $P^m(k,k-1)$ can be obtained by \myref{eq:pmxy2}, and the inequality can be derived by Bernoulli's inequality.
		
		2) For $\xi_t$ to take any population $X$ in $\chi-\chi^*$, the \textbf{point-wise drift} from $\xi_t$ to $\xi_{t+1}$ is derived by \myref{con:drift_t_without_d}:
		\begin{equation}
			\begin{aligned}
				\Delta(X) 
				= & 
				\mathbb{E}[d(\xi_t)-d(\xi_{t+1}) \mid \xi_t = X]
				\\ \leq &
				P(Y \in \chi_{(d_X-1)}\mid X)
			\end{aligned}%
			\label{con:drift_t_without_d}
		\end{equation}
		Thus, we have
		\begin{equation}
			\begin{aligned}
				\Delta(X \in \chi_d) &\leq P(Y \in  \chi_{d-1} \mid X \in \chi_d ) &\leq 
				\frac{d}{Q}
				\sum_{\gamma=1}^\lambda
				\gamma
				\frac{\pi_t(\chi_d^\gamma)}{\pi_t(\chi_d)}
			\end{aligned}
			\label{con:drift_t}
		\end{equation}
		where the last inequality can be obtained by formula \myref{equ:E(P(Y|X))_M2}. In the following, $\Delta(X \in \gamma_d)$ will be briefly denoted as $\Delta(d)$.
		
		3) Furthermore, the expectation of $\Delta(X)$ at $t$-th generation, i.e., the \textbf{average drift} $\bar {\Delta}_t$, can be calculated by \myref{con:ave_drift_t_M2},
		\begin{equation} 
			\begin{aligned}
				\bar {\Delta}_t 
				= & 
				\resizebox{1\width}{!}{  
					$ 
					\sum_{d=1}^n \Delta{(d)} {\pi_t{(\chi_d)}} /(1-\pi_t{(\chi^*)})$
				}
				\\ \leq &
				\frac{
					1
				}{
					Q
					(1-\pi_t(\chi^*))
				}
				\sum_{d=1}^n
				d
				\sum_{\gamma=1}^{\lambda}
				\gamma
				\pi_t(\chi_d^\gamma)
			\end{aligned}%
			\label{con:ave_drift_t_M2}
		\end{equation}
		
		4) Therefore, the EHT of the ENAS algorithm with \textbf{Mutation\#2} can be calculated by:
		\begin{equation} \nonumber
			\begin{aligned}
				\mathbb{E}[T(\xi_0)] = & {\mathbb{E}[d(\xi_0)]}/{\bar{\Delta}_t} 
				\\ \geq &
				\frac{
					Q
					(1-\pi_t(\chi^*))
					\sum_{d=1}^n
					d
					\pi_0(\chi_d)
				}{
					\sum_{d=1}^n
					d
					\sum_{\gamma=1}^\lambda
					\gamma
					\pi_t(\chi_d^\gamma)
				}
			\end{aligned}
		\end{equation}
		
	\end{IEEEproof}

	\begin{IEEEproof}[Proof of Theorem 4]
		The specific steps are as follows:
		
		1) At the $t$-th generation, $\xi_t$ can be any population $X$ in $\chi_k$. When $x\in X$ executes the bitwise mutation, its \textbf{mutation transition probability} is calculated as follows. 
		Once $X$ performs one step of mutation, the offspring population $Y$ belongs to $\chi$.
		Thus, the distance of $Y$ belongs to $[0,n]$. The probability of the population $Y$ belonging to $\chi_{k+D}$ can be expressed by~\myref{equ:E(P(Y|X))_M5}:
		\begin{equation}
			\label{equ:E(P(Y|X))_M5}
			\begin{aligned}
				& P(Y \in \chi_{k+D} \mid X \in \chi_k) 
				\\ = & 
				\prod_{x\in X} 
				\sum_{d_y\geq k+D}P^m(x,y) - 
				\prod_{x\in X} 
				\sum_{d_y\geq k+D+1} P^m(x,y)
			\end{aligned}
		\end{equation}
		where $D$ belongs to $[-k,n-k]$. The first term implies the probability that any offspring $y$ in $Y$ satisfies $d_y\geq k+D$, the second term means that any offspring $y$ in $Y$ satisfies $d_y>k+D$, and the subtraction of the two terms represents the probability that any one of the offspring in $Y$ satisfies $d_y\geq k+D$ and at least one offspring satisfies $d_y=k+D$.

		2) The \textbf{point-wise drift} can be calculated by \myref{con:drift_t_2_M5_without_d}:
		\begin{equation}
			\begin{aligned}
				\Delta(X) 
				= & 
				\mathbb{E}[d(\xi_t)-d(\xi_{t+1}) \mid \xi_t = X]
				\\ \leq &
				\sum_{d_Y=0}^{d_X-1}
				P(Y\in \chi_{d_Y}|X) \cdot (d_X-d_Y)
				\\ = &
				\sum_{D=1}^{d_X}
				P(Y\in \chi_{d_X-D}|X) \cdot D
			\end{aligned}
			\label{con:drift_t_2_M5_without_d}
		\end{equation}
		where the inequality is based on considering all $d_Y$ that can make $(d_X-d_Y)>0$. Thus, we have:
		\begin{equation}
			\begin{aligned}
				\Delta(X \in \chi_d) 
				\leq &
				\sum_{D=1}^{d}
				P(Y\in \chi_{d-D}|X\in \chi_d) \cdot D
				\\ = &
				d \cdot 
				\prod_{x\in X} \sum_{d_y\geq 0} P^m(x,y)
				\\  & -
				\sum_{D=0}^{d-1} 
				(
				\prod_{x\in X}  \sum_{d_y\geq d-D} P^m(x,y)
				)
				\\ \leq &
				d - 
				\sum_{D=1}^{d} 
				(
				\mathop{\min}\limits_{d_x \in [d,n]}
				\{
				\sum_{d_y\geq D} P^m(x,y)
				\}
				)^\lambda
			\end{aligned}
			\label{con:drift_t_2_M5}
		\end{equation}
		where $P^m(x,y)$ denotes the bitwise mutation transition probability of individual $x$ to $y$, and the last inequality is derived by considering for any $x\in X$ there is $\sum_{d_y\geq 0}P^m(x,y)=1$.
		
		3) Furthermore, the \textbf{average drift} at the $t$-th generation can be calculated by:
		\begin{equation} 
			\begin{aligned}
				\bar {\Delta}_t 
				\leq 
				\sum_{d=1}^n 
				\frac{
					\pi_t{(\chi_d)}
				}{1-\pi_t{(\chi^*)}}
				(
				d - 
				\sum_{D=1}^{d} 
				(
				\mathop{\min}\limits_{d_x \in [d,n]}
				\{
				\sum_{d_y\geq D} P^m(x,y)
				\}
				)^\lambda
				)
			\end{aligned}
			\label{con:ave_drift_t_M5}
		\end{equation}
		
		4) According to the \textbf{Mutation\#4} transition probability, the EHT of the ENAS algorithm is derived by:
		\begin{equation} \nonumber
			\begin{aligned}
				&\mathbb{E}[T(\xi_0)] = {\mathbb{E}[d(\xi_0)]}/{\bar{\Delta}_t} 
				\\ \geq &
				\frac{
					\resizebox{1\width}{!}{
						$(1-\pi_t{(\chi^*)})
						\sum_{d=1}^n d \pi_0{(\chi_d)}$
					}
				}{
					\resizebox{0.9\width}{!}{  
						$
						\sum_{d=1}^{n}
						(
						d-
						\sum_{D=1}^{d}
						(
						\mathop{\min}\limits_{d_x \in [d,n]}
						\{
						\sum_{d_y\geq D}
						P_4^m(x,y)
						\}
						)^\lambda
						)
						\cdot
						\pi_t{(\chi_d)}
						$
					}
				}
			\end{aligned}
		\end{equation}
		
	\end{IEEEproof}